\documentclass[aps,reprint]{revtex4-2}

\usepackage{graphicx}%
\usepackage{dcolumn}%
\usepackage{bm}%
\usepackage{tabularray}
\usepackage{makecell}
\usepackage{float}
\usepackage{placeins}
\usepackage{caption}
\usepackage{subcaption}
\usepackage{booktabs}
\usepackage{listings}
\usepackage{xspace}
\usepackage{xcolor}

\usepackage{amsmath, amsthm, amssymb}

\usepackage{hyperref}       

\usepackage{cleveref}

\newtheorem{prop}{Proposition}

\newcommand{\methodname}{Transformers\xspace}

\def\eqref#1{equation~\ref{#1}}

\def\1{\bm{1}}

\def\vr{{\bm{r}}}

\DeclareMathAlphabet{\mathsfit}{\encodingdefault}{\sfdefault}{m}{sl}
\SetMathAlphabet{\mathsfit}{bold}{\encodingdefault}{\sfdefault}{bx}{n}

\newcommand{\R}{\mathbb{R}}

\newcommand{\update}[1]{{#1}}

\begin{document}

\preprint{APS/123-QED}

\title{\textbf{Transformers Discover Molecular Structure Without Physical Priors} 
}%

\author{Tobias Kreiman$^1$}
\author{Yutong Bai$^1$}%
\author{Fadi Atieh$^1$}
\author{Elizabeth Weaver$^1$}
\author{Eric Qu$^1$}
\author{Aditi S. Krishnapriyan$^{1,2}$}
\affiliation{Email: \href{mailto:tkreiman@berkeley.edu}{tkreiman@berkeley.edu}, \href{mailto:aditik1@berkeley.edu}{aditik1@berkeley.edu} \\ $^1$UC Berkeley, $^2$LBNL}

\begin{abstract}

Computational simulations play a central role in scientific discovery, and machine learning (ML) has emerged as a promising alternative to traditional physics-based modeling. However, scientific modeling requires physically meaningful predictions, raising a fundamental question for data-driven methods: to what extent can physical inductive biases---that is, prior assumptions about the structure of the physical world---emerge by learning from data alone? In this work, we study atomistic modeling, a representative field in the computational sciences where ML architectures have historically embedded strong physical inductive biases---such as geometric locality and graph structure. We develop a systematic understanding of how physical patterns can alternatively be discovered directly from data by training a model without domain-specific priors, including any manually defined atomistic pairwise interactions. We find that the model autonomously recovers key physical structure, such as \update{learned interatomic interaction strengths} that mirror classical electrostatics and interaction cutoffs consistent with traditional physical models. We further demonstrate predictable neural scaling law behavior with increased data and compute, and find accuracy on certain metrics competitive with physics-informed architectures. Our results clarify the boundary between engineered inductive biases and learnable physical structure, suggesting that general-purpose architectures can serve as principled baselines for scientific modeling by learning fundamental physical structure directly from data.

\end{abstract}

\maketitle

Computational modeling is an essential tool for studying the properties of physical systems. While \textit{ab initio} methods build on known physical laws to accurately describe the behavior of matter, their high computational cost often makes them prohibitively expensive for systems of practical interest. Machine learning (ML) methods have emerged as a promising alternative, leveraging data to model physical systems at a fraction of the cost of traditional approaches. However, effectively applying ML in the physical sciences requires careful consideration to ensure physically meaningful predictions. Consequently, many ML approaches for problems in the physical sciences incorporate physical priors, such as specific symmetries, directly into domain-specific model architectures \citep{batatia_mace_2022, Batzner_2022nequip, zhang2026graphneuralnetworkera, gasteiger_gemnet_2021, Jumper2021, li2021fourierneuraloperatorparametric, chang2025spacegroupequivariantcrystal, krishnan2024hamiltonianbridgephysicsdrivengenerative, Behler2007}.  

In contrast, the broader ML landscape has shifted toward a domain-agnostic paradigm centered on the Transformer architecture~\citep{dosovitskiy2021imageworth16x16wordsvit, vaswani2023attentionneed, octomodelteam2024octoopensourcegeneralistrobot, brown2020languagemodelsfewshotlearners, touvron2023llamaopenefficientfoundation}. The success of the Transformer has been guided by empirical scaling laws that allow for the precise forecasting of performance gains as a function of dataset size, computational budget, and number of model parameters \citep{kaplan2020scaling, hoffmann2022trainingcomputeoptimallargelanguage}. This common architectural framework, paired with established scaling laws, has dramatically accelerated the research process by providing a standardized recipe for addressing a broad class of ML problems. As a result, this framework has enabled the development of powerful multi-modal models \citep{grattafiori2024llama3herdmodels, octomodelteam2024octoopensourcegeneralistrobot}, specialized hardware and software for Transformers \citep{pytorch2, jouppi2017tpu, kwon2023efficientmemorymanagementlarge}, and beyond. 

The appeal of data-driven approaches for the physical sciences paired with the success of domain-agnostic methods in deep learning raise a fundamental question: to what extent must physical laws be explicitly encoded into ML models, and to what extent can they be learned directly from data? This tension is most apparent for domain-agnostic architectures such as Transformers, which have unified tasks across language, vision, and robotics without explicit domain-specific inductive biases. Despite this, their adoption in the physical sciences has been limited, partly due to the historically smaller availability of data and the resulting perspective that models without physical inductive biases cannot recover complex physical interactions without explicit geometric guidance \citep{batatia2024macemp, Batzner_2022nequip, gasteiger_gemnet_2021, de_Silva_2020, lemos2022rediscoveringorbitalmechanicsmachine, brehmer2025doesequivariancematterscale, Behler2007}. However, the recent emergence of massive scientific datasets, such as Open Molecules 2025 (OMol25) \citep{levine2025openmolecules2025omol25}, provides a unique opportunity to test the limits of this assumption using atomistic modeling as a representative domain of computational modeling with rich physical structure.

In this work, we investigate how the fundamental principles of molecular interaction can be autonomously discovered from data. In contrast to the prevailing paradigm, which prescribes locality through an interaction graph and encodes geometric features directly into model architectures \citep{Batzner_2022nequip, batatia_mace_2022, gasteiger_gemnet_2021, schutt_schnet_2017, Behler2007}, we instead represent a molecular system simply as a set of atomic coordinates in Cartesian space. Without constructing a graph or embedding chemical priors, we investigate whether an unmodified Transformer can recover fundamental physical structure when trained to predict molecular energies and forces. Probing the model’s internal representations, we find that the Transformer autonomously recovers functional forms consistent with classical physics, such as \update{a learned interatomic interaction} strength that mirrors electrostatic interactions. Transformers also learn to \textit{flexibly adapt their effective receptive field}, \update{focusing on long-range interactions in less dense regions and concentrating on local ones for tightly packed atoms}. Interestingly, the learned effective cutoff matches the cutoff used in hand-designed physics-based ML models \citep{Batzner_2022nequip, gasteiger_gemnet_2021}. We also observe that these physically consistent behaviors emerge even in low-data regimes and that performance continues to predictably improve with training resources.

While it remains crucial to assess the failure modes of ML models for the physical sciences before broad application to scientific discovery \citep{chiang2025mlip, deng2024overcomingsystematicsofteninguniversal, dengmptrj, kreiman2025understandingmitigatingdistributionshifts, fu2023forces, bihani2023egraffbenchevaluationequivariantgraph}, our experiments suggest that, given the large chemical datasets now available \citep{levine2025openmolecules2025omol25, Chanussot_2021_oc20, barrosoluque2024openmaterials2024omat24, dengmptrj}, many physical inductive biases can be learned directly from data. These results pave the way for transferring insights from the broader ML literature into the atomistic modeling community, leveraging a general and flexible architecture capable of addressing a wide range of problems \citep{yuan2025foundationmodelsatomisticsimulation}.

\paragraph*{What can Transformers Learn without Physical Priors?}

We investigate how Transformers learn physical inductive biases by training a model without built-in physical priors to predict molecular energies and forces (see \Cref{apx: md_transformer} for details on the model). The model simply maps raw cartesian coordinates and atomic types, along with tokens encoding global information like charge and spin, to a molecule-level energy and per-atom forces. The key operation in the Transformer is the self-attention mechanism that transforms a hidden state $X \in \R^{n\times d}$ of $n$ elements with dimension $d$ by applying:
\begin{equation}
\label{eqn:multi_head_attention}
X \leftarrow X + \text{softmax}\left[(XW_Q)(XW_K)^T\right] XW_V,  
\end{equation}
where the matrices $W_Q, W_K, W_V$ are learnable parameters and the softmax applies to each row of $(XW_Q)(XW_K)^T\in \R^{n\times n}$ to yield normalized \textit{attention scores}. \update{These attention scores measure how strongly one atom influences the representation of another, providing a measure of the model's \emph{learned interaction strength} between atoms.} This setup ensures that any physical structure in the model---including any learned pairwise relationships reflected in the attention---at the end of training arises from discovering patterns in the data, not knowledge built into the model architecture.  

We ground our study in the recently released OMol25 dataset \citep{levine2025openmolecules2025omol25}, which provides energy and force labels calculated at the $\omega$B97M-V/def2-TZVPD level of theory. OMol25 covers a broad range of chemical structures, from biomolecules to metal complexes to electrolytes. The diversity and abundance of data makes it an ideal place to study molecular representation learning \textit{without} graph priors. We simply train our model to predict the energy and force labels in the OMol25 dataset using a two-staged training pipeline that first pre-trains to predict a discretized version of the targets and then fine-tunes on the continuous values. We describe our training procedure in further detail in \Cref{sec:md_and_sl}. 

Once trained, we first analyze the learned representations of our graph‑free model to understand what physical information it captures. Next, we examine how Transformer performance scales with data and compute on both molecular and synthetic data to explore the robustness of the learned physical patterns under different training resources.

\begin{figure}
    \centering
    \includegraphics[width=0.9\linewidth]{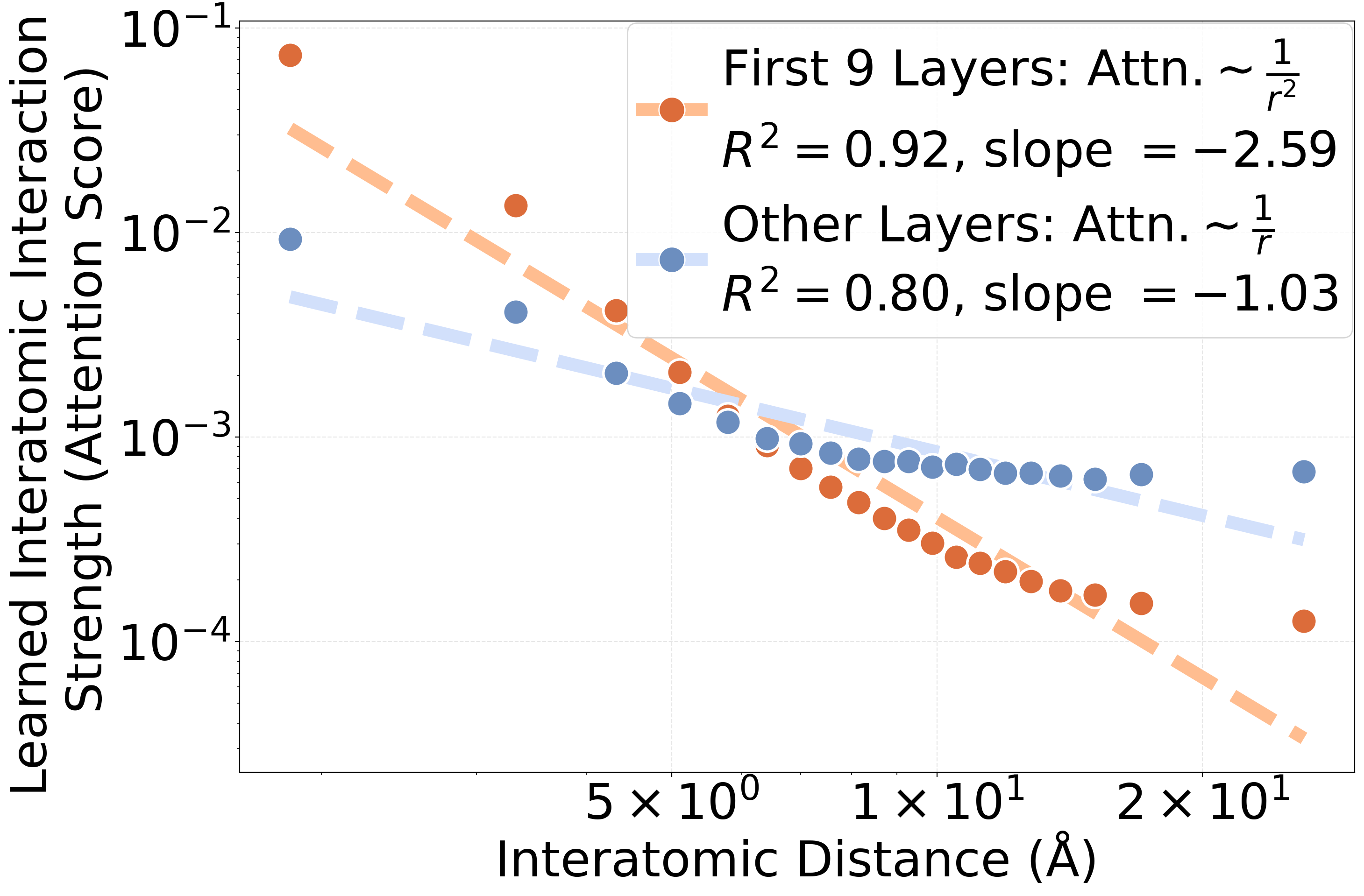}
    \caption{\update{\textbf{Transformers learn interatomic interaction strengths consistent with physical models.} Averaged across validation examples in OMol25, we plot the model's attention scores as a function of interatomic distance. The attention scores quantify how strongly the representation of one atom is updated using information from another, and thus quantifies the model's learned interaction strength. Each dot is the mean attention score within an interatomic distance quantile. Attention decays sharply with interatomic distance in early layers, following a trend closer to the $1/r^2$ scaling associated with electrostatic forces. In later layers, the decay is more gradual, resembling the $1/r$ scaling of electrostatic energies, suggesting the model learns local features before focusing on global ones in deeper layers. At larger distances ($\gtrsim 20$ Å), the attention profile flattens, likely reflecting the scarcity of atom pairs at such separations in the training data.}}
    \label{fig:attn_v_distance}
    \vspace{-6pt}
\end{figure}

\paragraph*{Investigating the Learned \update{Interatomic Interaction Strength} of the Transformer.}
\update{To study the model's learned interatomic interaction strength, we focus on the learned \textit{attention scores}---the softmax-normalized dot products between queries and keys---which reveal how the Transformer models pairwise relationships between atoms (see \Cref{eqn:multi_head_attention}).} When we plot the models learned interaction strength (attention score) against interatomic distance (see \Cref{fig:attn_v_distance}), we observe a clear inverse correlation. In early layers, the attention scores drop off sharply with an exponent $\sim -2$, resembling the decay of electrostatic forces. In later layers, the drop-off in attention score is more gradual with an exponent of $\sim -1$, more in line with the decay of energies. Across all layers, the attention pattern flattens off at large interatomic distances ($\gtrsim 20$ Å), likely due to the lack of atom pairs at such distances in the dataset (relative to the number of atoms that are closer together).

\update{One interpretation of these results is that the Transformer learns to first model local interactions before shifting to global ones. In early layers, the model concentrates on nearby atoms, evident from the learned $\sim 1/r^2$ interaction strength (shown in \Cref{fig:attn_v_distance})}. Interestingly, this distance-dependent attention drops off around the $6$–$12$ Å range, coinciding with the radius cutoffs commonly used in traditional graph-based ML models for atomistic systems  \citep{kovács2023maceoff23, fu2025learningsmoothexpressiveinteratomic_esen, levine2025openmolecules2025omol25}. \update{In later layers, the model shifts toward long-range interactions, as evidenced by an interatomic interaction strength that scales as $\sim 1/r$. These deeper layers also focus increasingly on global molecular properties, such as charge and spin, compared to the initial layers (see \Cref{fig:attn_by_type}). These findings suggest that the Transformer learns a physically meaningful model directly from the data. It first extracts local features in the early layers---without relying on predefined graph constructions or message-passing schemes---and subsequently refines the representation using long-range interactions and global electronic states.}

\paragraph*{Investigating adaptive interaction patterns.} \update{We observed a clear decay of interaction strength, as represented by attention scores, with interatomic distance; the Transformer learns this pattern directly from data without relying on a predefined graph structure.} This prompts us to explore whether this decay pattern is adaptive. In contrast, graph neural networks (GNNs) often use a fixed radius cutoff to determine which pairs of atoms are considered neighbors. However, this fixed radius may be optimal for one molecule but not for another~\citep{kreiman2025understandingmitigatingdistributionshifts, digiovanni2023oversquashingmessagepassingneural, dwivedi2023longrangegraphbenchmark}. Furthermore, different radii might even be warranted for different atoms \textit{within the same molecule}. For example, if an atom is in a tightly packed region with many neighboring atoms, short-range interactions might overwhelm long-range interactions. However, if the atom is isolated from the rest of the molecule, then long-range interactions would increase in relevance.

\update{To investigate whether the Transformer learns these flexible patterns, we define the model's learned \textit{effective interaction radius} $R_i$
for atom $i$ as follows}:

\begin{equation}
\label{eqn:effective_r}
\vspace{-2pt}
    R_i = \inf \Biggl\{ R \in \mathbb{R}_{\geq 0} : \sum_{\substack{j:\\ \|\vr_i-\vr_j\|_2 \le R}} a_{ij} \geq \delta \Biggr\},
    \vspace{-2pt}
\end{equation}
\update{where $\vr_i$ is the position of atom $i$ and $a_{ij}$ is the attention score between atom $i$ and $j$. This radius is the smallest distance that contains $\delta=90\%$ of atom $i$'s total attention. We plot this radius against a simple measure of local sparsity: the median distance to all other atoms in the structure. For reference, we compute two pairwise proxies that characterize short- and long-range interactions from the raw DFT output files. The short-range reference is based on the off-diagonal atomic blocks of the L\"owdin-orthogonalized one-particle density matrix, which measures electronic delocalization between atoms and is expected to be nearsighted \citep{Wiberg1968ApplicationOT, Mayer1983ChargeBO, Prodan2005}. The long-range reference is a monopole electrostatic proxy computed from L\"owdin partial charges. We find that early model layers learn radii that are nearly independent of neighborhood sparsity and align with the density-matrix reference, whereas later layers learn an adaptive interaction cutoff that expands in sparse environments and tracks the electrostatic reference (see \Cref{fig:attn_effective_radius} and \Cref{apx: more_reps} for details).}

Finally, we observe that specific attention heads learn other varied flexible interaction patterns that are difficult to anticipate or hard-code \textit{a priori}. For example, we observe certain heads exhibit a $k$-nearest-neighbor behavior (kNN), where attention is directed based on the rank of neighboring atoms rather than simply decreasing with distance. Other heads show non-monotonic attention patterns, and some even increase attention with distance. For further visualizations, see the \Cref{fig: special_knn_behavior} and \Cref{fig: special_dist_behavior}. 

\paragraph*{Observations.} The analysis in this sections reveals that simple ML models can recover complicated structure. Importantly, the learned structure is physically meaningful, with learned cutoffs that match the hand-picked values in current physics-based ML models \citep{Batzner_2022nequip, gasteiger_gemnet_2021} and patterns that align with physical models. Paired with the diverse, interpretable head behavior, these results support the hypothesis that many inductive biases, traditionally incorporated directly into model design, can be discovered from data directly. 

\paragraph*{Neural Scaling Analysis}

\begin{figure}
    \centering
    
    \begin{subfigure}[t]{0.3\linewidth}
        \centering
        \includegraphics[width=0.95\linewidth]{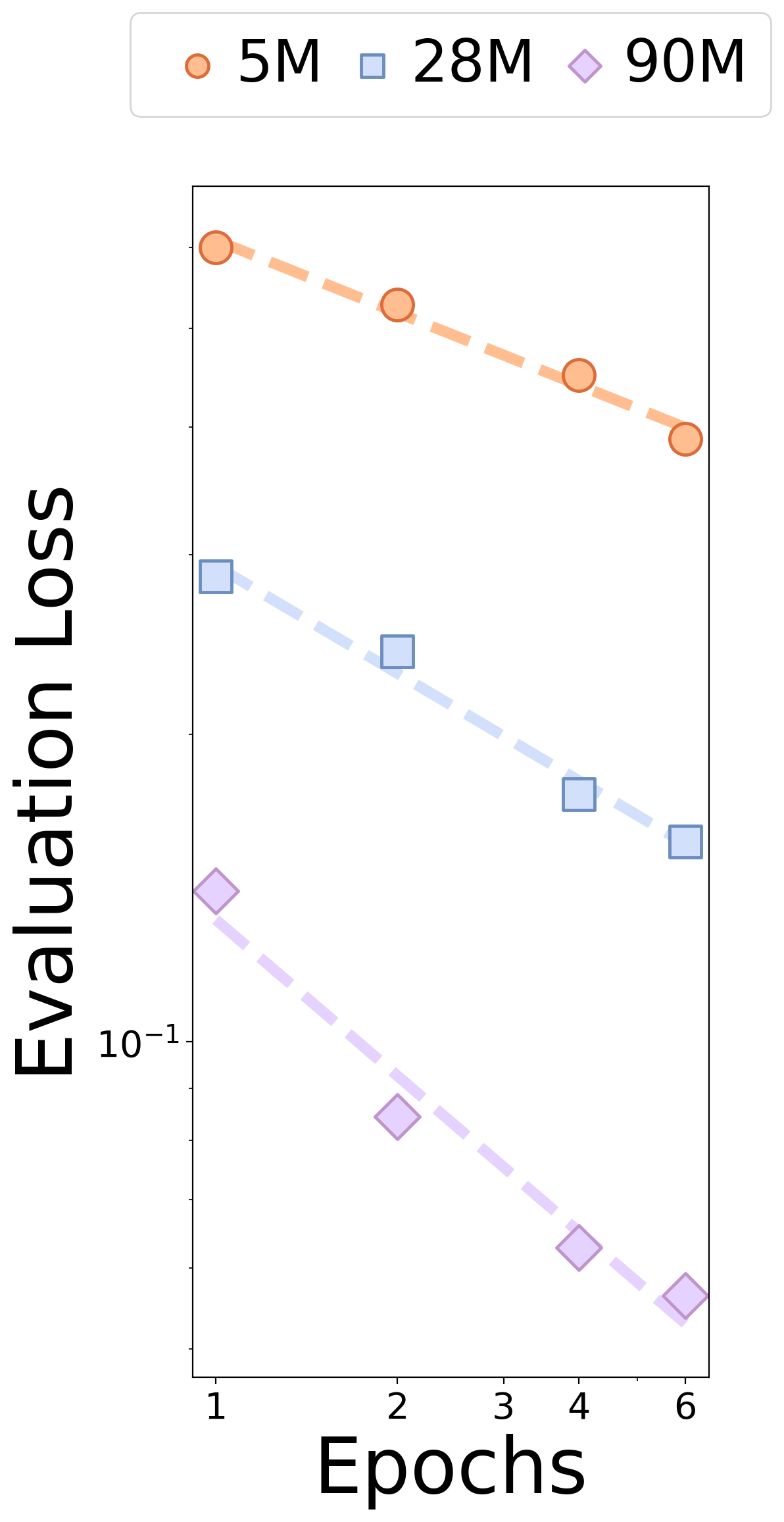}
        \caption{\textbf{Data Scaling.}}
        \label{fig:ft_data_scaling}
    \end{subfigure}%
    \begin{subfigure}[t]{0.75\linewidth}
        \centering
        \includegraphics[width=0.92\linewidth]{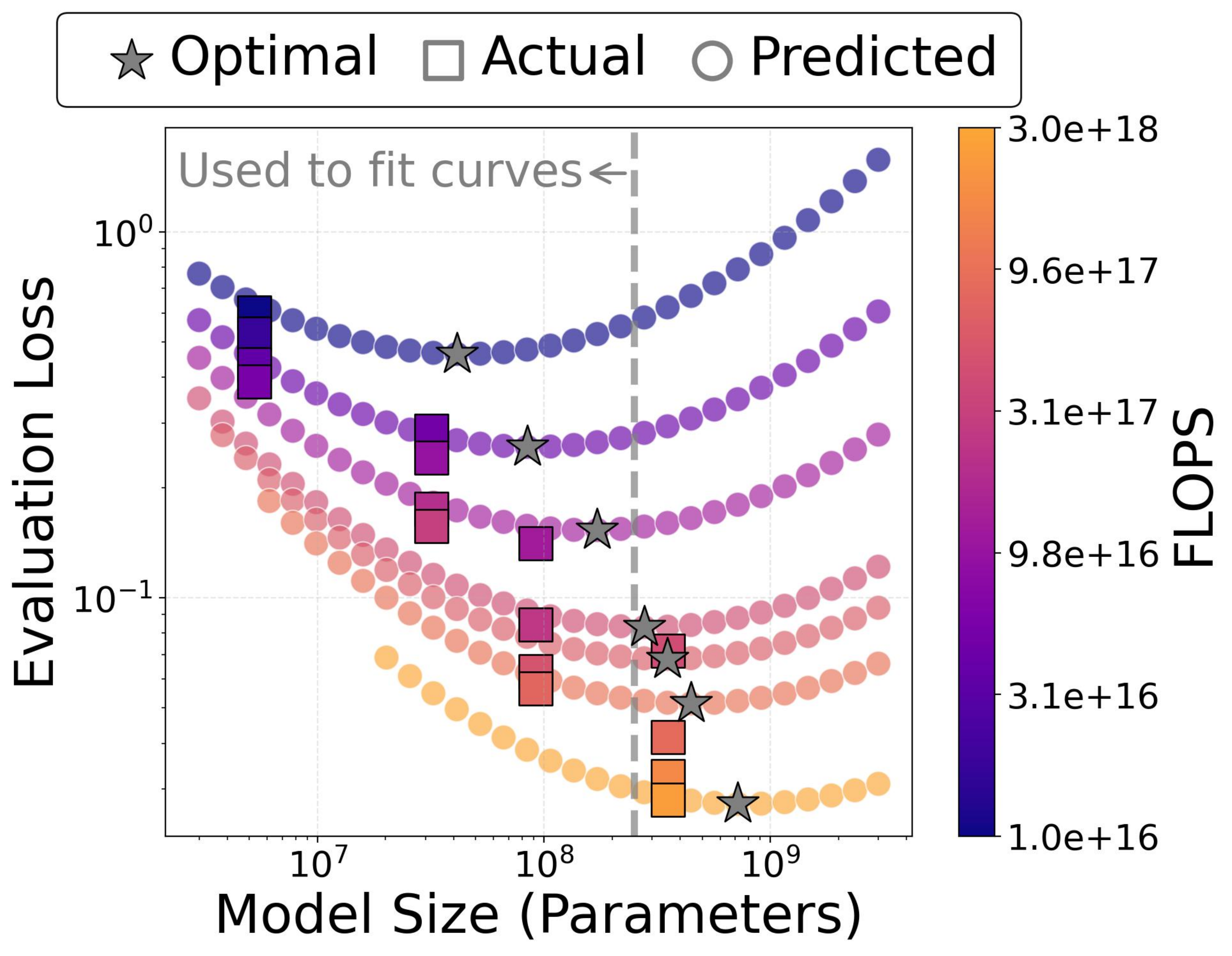}
        \caption{\textbf{Fine-Tuning IsoFLOP Curves.}}
        \label{fig:ft_flops}
    \end{subfigure}%
    \caption{\textbf{\methodname scale predictably with training resources when modeling molecules.} On the OMol25 4M dataset, we train Transformer models with 5M, 30M, 90M, and 306M parameters for different numbers of epochs while holding all other training hyperparameters fixed. Evaluation performance follows a clear power-law relationship with model size. (a) We fit scaling laws using the three smaller models and accurately predict the performance of larger models. (b) We plot predicted IsoFLOP curves---curves of constant training compute measured in floating-point operations (FLOPs)---and find that they are parabolic, with the optimal model size and achievable performance following consistent power-law relationships, in agreement with previous work~\citep{hoffmann2022trainingcomputeoptimallargelanguage}. \update{Scaling laws provide a complementary framework to physics-based ML model design by predicting how model performance improves with increasing training resources.}}
    \label{fig:scaling_laws_omol}
    \vspace{-6pt}
\end{figure}

To confirm whether the Transformer consistently learns physically meaningful structure and understand how performance depends on training resources, we perform a scaling law analyses. Scaling laws describe how model performance improves predictably with training resources. They are widely used in other areas of ML to guide model design and training with predictable results~\citep{hoffmann2022trainingcomputeoptimallargelanguage, kaplan2020scaling, brown2020languagemodelsfewshotlearners, grattafiori2024llama3herdmodels, snell2024scalingllmtesttimecompute}. \update{These scaling laws complement traditional physics-based ML modeling. While physics-based approaches encode physical laws directly into model architectures, scaling laws provide predictable behavior by empirically characterizing how model capabilities improve with additional training resources. Together, they provide complementary tools for designing more capable molecular models.}

\paragraph*{Fine-tuning scaling laws.} We train three model sizes (5M, 30M, 90M) for varying numbers of epochs (1, 2, 4, 6) on OMol25 4 million \citep{levine2025openmolecules2025omol25}. All runs use identical hyperparameters, with only the parameter count and training epochs varied. Following prior work \citep{kaplan2020scaling, hoffmann2022trainingcomputeoptimallargelanguage}, we assume a power-law relationship between the validation loss $\mathcal{L}$ and model size $N$: $\mathcal{L}(N) = \left( \frac{N}{N_c} \right)^\alpha,$ where $N$ represents the number of non-embedding parameters. We fit power-law scaling curves that predict model performance from both model size and number of training epochs (see \Cref{fig:ft_data_scaling}). From these runs, we generate IsoFLOP curves---performance curves for constant training compute (total floating-point operations). On each curve, smaller models are trained longer, while larger models are trained for fewer epochs. The predicted IsoFLOP curves are parabolic, with an optimal model size for each compute budget that follows a clear power-law relationship, consistent with findings in other domains \citep{hoffmann2022trainingcomputeoptimallargelanguage}. The scaling laws accurately predict the performance of a larger 306M parameter model, which was not used in fitting the curves (see \Cref{fig:ft_flops}). We also report pre-training scaling laws in \Cref{fig:pt_model_scale} for models up to 1B parameters. 

\paragraph*{Observations.} These results show that Transformers for molecular modeling scale predictably with training resources in both pre-training and fine-tuning. Given that performance has not saturated at 1B parameters, and that scaling laws in other fields hold up to much larger model sizes~\citep{touvron2023llama2openfoundation, brown2020languagemodelsfewshotlearners, siméoni2025dinov3}, it is plausible that accuracy could continue to improve well into the hundreds of billions of parameters. While scaling laws can break down in certain modalities~\citep{henighan2020scalinglawsautoregressivegenerative}, our findings suggest that molecular Transformers may follow the same predictable trends that have driven progress in other fields of ML.

\begin{figure}
    \centering
    \includegraphics[width=0.8\linewidth]{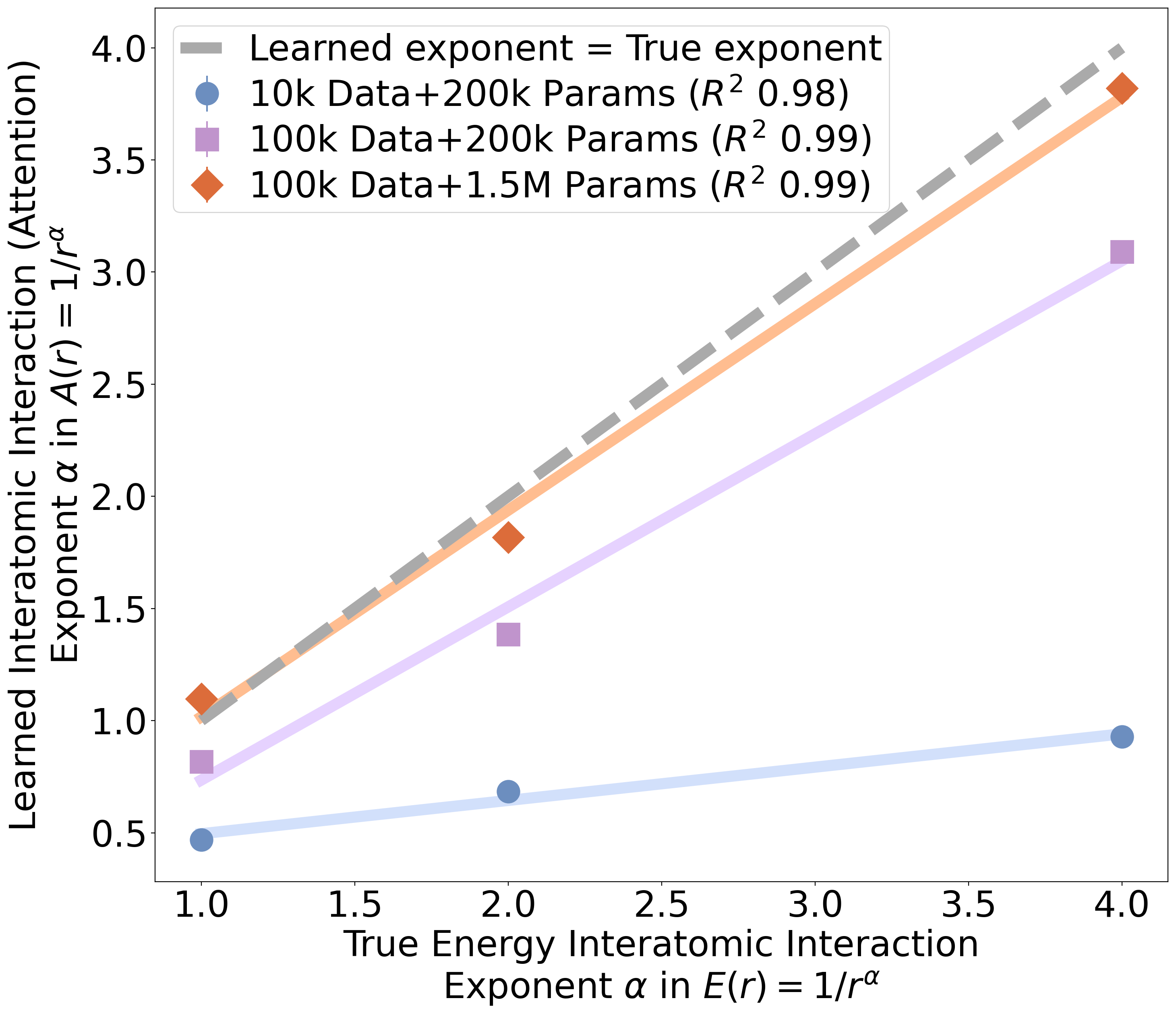}
    \caption{\textbf{Transformers learn interatomic interaction energies in their attention scores and become more accurate as data and model size are scaled.} We generate three synthetic datasets with pairwise potentials of the form $ E(r) = 1 / r^{\alpha}$ for $\alpha \in \{1, 2, 4\}$, and train a Transformer to predict the total energy. \update{We estimate the model's learned interatomic interaction strength by fitting $\alpha$ in $A(r) \propto 1/r^\alpha$, where $A(r)$ is the model's learned attention score for atoms $r$ apart.} In the low-data, small-model regime, the learned interaction exponent is strongly correlated with the true exponent used to generate the data ($R^2 = 0.98$), but its absolute value deviates substantially from the ground truth. As data and model size are increased, the learned interaction exponent increasingly matches the true underlying exponent. All points plotted here are averaged over five seeds (the standard deviation is $<0.02$ for all points).
}
    \label{fig:synthetic}
    \vspace{-6pt}
\end{figure}

\paragraph*{Synthetic data scaling laws.} \update{To further probe how Transformers learn physical concepts, we perform an experiment on synthetic data where we can control the energy labels for each atomistic structure. Specifically, we probe how the model's learned interaction strength aligns with the true physical potential as a function of data and model size.} We generate three synthetic datasets with energy labels defined by a simple pairwise potential. To ensure that the molecular structures are physically meaningful, we sample geometries from the OMol25 dataset and assign each structure a total energy given by:
\begin{equation}
\vspace{-2pt}
\label{eqn:synthetic_energy}
E(\vr) =
\sum_{\substack{1 \le i,j \le n \\ i \neq j}}
\frac{1}{\| \vr_i - \vr_j \|_2^{\alpha}},
\vspace{-2pt}
\end{equation}
where $n$ is the number of atoms in the molecule and $\alpha \in \{1, 2, 4\}$.

For each dataset, we train a Transformer to predict the total energy of each system. Using a held-out validation set also sampled from OMol25, \update{we estimate the model's learned interatomic interaction exponent by fitting the relationship $A(r) \propto 1/r^\alpha$, where $A(r)$ is the model's learned attention score for atoms separated by a distance $r$.} \Cref{fig:synthetic} shows that the \update{learned interaction exponent correlates strongly with the true potential exponent} used to generate the energy labels ($R^2 = 0.98$), although its absolute value deviates from the ground truth in the low-data regime. As data and model size are increased, the learned exponent increasingly approximates the true exponent of the underlying energy function. We emphasize that the data and model scales used in this experiment are small ($\sim1,000\times$ less data than the full OMol25 dataset), yet the model still recovers the $1 / r^{\alpha}$ structure in its attention scores. We observe that the Transformer slightly underpredicts the true exponent, particularly for the larger exponents. One possible explanation is that the number of atomic pairs at large separations is small, which causes the attention scores to decay more slowly with distance at large $r$ (also observed in \Cref{fig:attn_v_distance}).

To conclude our investigation with the synthetic data, we outline a construction that provides intuition for how Transformers can learn this fundamental $1/r^{\alpha}$ relationship. We construct query and key matrices that induce Gaussian kernels over interatomic distances. Mixtures of these kernels can then approximate inverse-power-law relationships of the form $1 / r^{\alpha}$. The details of this construction are provided in \Cref{apx:construction}. We emphasize that this argument is not intended as a fully rigorous proof, but rather as an illustration that builds intuition for how Transformers can learn physically meaningful functions.

\paragraph*{Further evaluations.} \update{Beyond learned interatomic interaction patterns, we analyze a water dimer to evaluate \emph{intermolecular} interactions. \Cref{fig:water_dimer} shows that both the Transformer and eSEN-sm (a symmetry-constrained GNN), trained on OMol25, accurately reproduce the interaction energy curve in agreement with DFT and experimental measurements \citep{RocherCasterline2011}. The Transformer implicitly learns a smooth decay in the interaction energy, avoiding the abrupt truncation of interactions caused by a fixed radius cutoff.}

\begin{figure}
    \centering
    \includegraphics[width=0.85\linewidth]{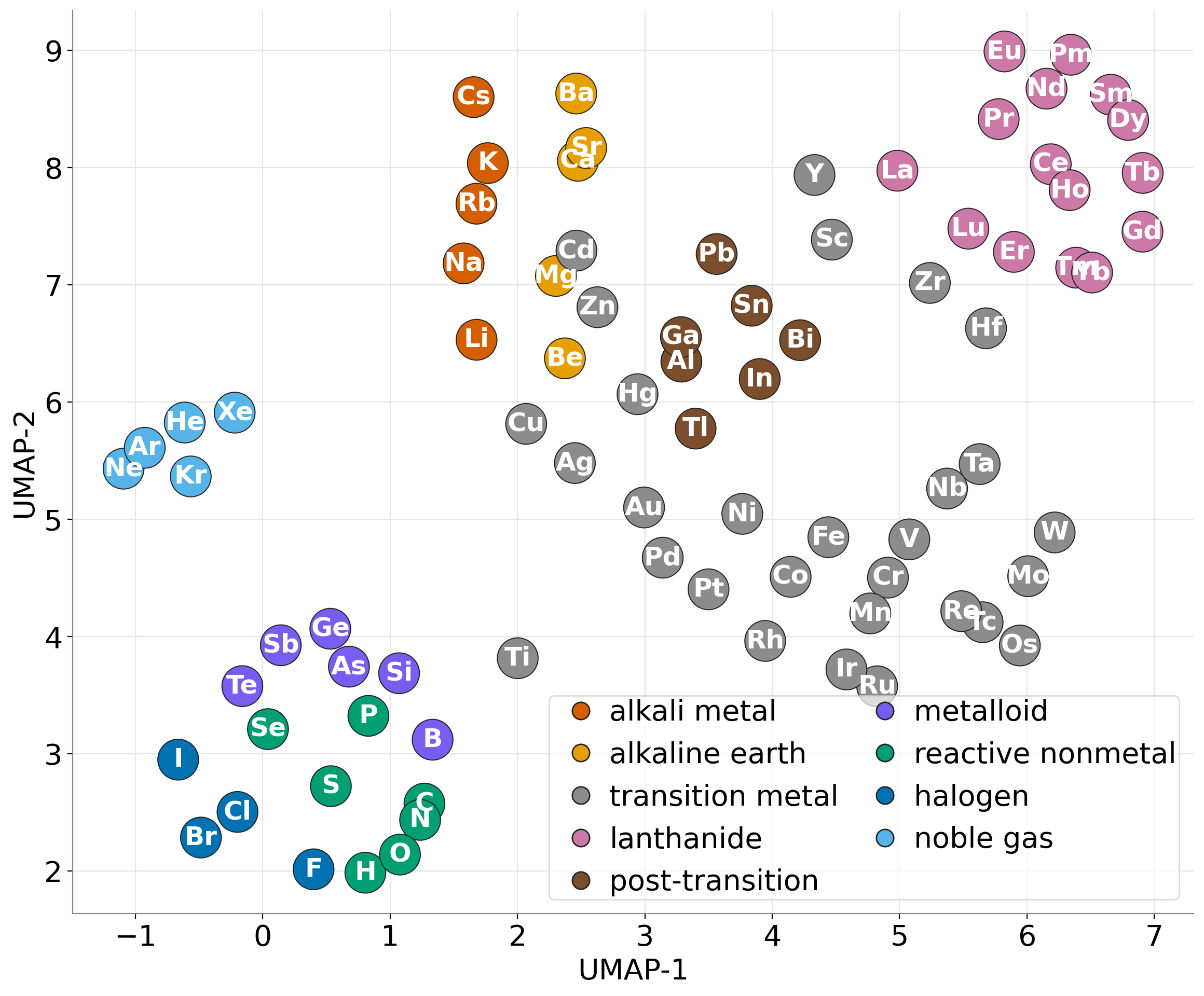}
    \caption{\update{\textbf{Transformers learn chemically meaningful relationships between elements.} We use UMAP to visualize the Transformer's learned elemental embeddings, which recover trends in the periodic table. For example, noble gases form a distinct cluster, while the alkali and alkaline earth metals follow their periodic ordering.}}
    \label{fig:elem_umap}
    \vspace{-6pt}
\end{figure}

\update{We next investigate the model's learned elemental representations. \Cref{fig:elem_umap} visualizes the learned element embeddings using UMAP \citep{mcinnes2020umapuniformmanifoldapproximation}. The Transformer organizes the elements into a highly structured representation that reflects the periodic table: noble gases form a distinct cluster, while the alkali and alkaline earth metals follow their correct ordering by period. Although UMAP provides only a low-dimensional projection, these results suggest that the model captures fundamental relationships between elements.}

We finally examine performance on the quantum mechanical energy and force prediction task. Perhaps most surprisingly, when we follow the scaling laws and train a 1B parameter Transformer, we find that it also achieves competitive accuracy on the energy and force prediction task compared to a physics-based GNN \citep{fu2025learningsmoothexpressiveinteratomic_esen} \textit{under a matched training compute budget and using the same training dataset}. We note that we evaluate the energy and force errors on out-of-distribution molecular compositions, suggesting that the model has some robustness in its learned physics. For further evaluations, \update{including molecular dynamics simulations}, see \Cref{apx: further_exp}.

\paragraph*{Conclusion.}
We investigated the learned representations of an unmodified, graph-free Transformer trained on chemical data. We found that the Transformer naturally learns physically consistent behaviors and that its performance improves predictably with training resources. Interestingly, since the Transformer includes no explicit physical inductive biases, we found that it exhibits adaptive patterns---such as an effective radius cutoff that varies based on atomic environments---which could be hard to specify \textit{a priori}.

More generally, insights from previous deep learning research suggest that when enough data is available, expressive models that leverage powerful optimization algorithms on modern hardware can outperform methods that rely on hand-crafted inductive biases \citep{brown2020languagemodelsfewshotlearners, vaswani2023attentionneed, dosovitskiy2021imageworth16x16wordsvit, kim2024openvlaopensourcevisionlanguageactionmodel}. While some inductive biases may be beneficial for specific problem settings, tackling new problems often requires designing new biases for each task. In contrast, general search and learning methods can discover inductive biases directly from data, and perhaps even discover more flexible solutions that are hard to anticipate \textit{a priori} \citep{Sutton}.

Our experiments suggest that Transformers appear capable of learning many of the physical inductive biases typically incorporated in current ML models for atomistic modeling. We hope these findings point towards a standardized, widely applicable architecture for atomistic modeling that draws on insights from the broader deep learning community.

\begin{acknowledgments}

We would like to thank Yossi Gandelsman, Sanjeev Raja, Ritwik Gupta, Alyosha Efros, Rasmus Malik Hoeegh Lindrup, and the ASK lab for the fruitful discussion and feedback. This work was supported by the Toyota
Research Institute as part of the Synthesis Advanced Research Challenge. This work was also partially supported by Scialog grant \#SA-AUT-2024-015b from Research Corporation for Science
Advancement and Arnold and Mabel Beckman Foundation. This research used resources of
the National Energy Research Scientific Computing Center (NERSC), a U.S. Department of Energy Office
of Science User Facility located at Lawrence Berkeley National Laboratory, operated under Contract No.
DE-AC02-05CH11231. This work used DeltaAI at the National Center for Supercomputing Applications through allocation  CIS250587 from the Advanced Cyberinfrastructure Coordination Ecosystem: Services \& Support (ACCESS) program, which is supported by National Science Foundation grants \#2138259, \#2138286, \#2138307, \#2137603, and \#2138296.
\end{acknowledgments}

\bibliography{references}%

\newpage
\appendix
\section{Extended Related Work}
\label{sec:related}

\paragraph*{Machine learning atomistic property prediction.} Machine learning (ML) atomistic property predictors learn to map chemical representations, like atomic types and positions, to physical properties, such as the energy of the system. 
ML atomistic property predictors typically incorporate a number of physical inductive biases through graph-based architectures \citep{batatia_mace_2022, Batzner_2022nequip, gilmer2017neural, gasteiger_gemnet_2021}. Recent models have started to discard some of the hard symmetry constraints used by previous models \citep{qu2024importance, neumann2024orbfastscalableneural, mazitov2025petmadlightweightuniversalinteratomic}, but they continue to rely on graph neural networks (GNNs) as the architectural backbone. While some GNNs do incorporate attention-based mechanisms \citep{gat_oversmooth, liao2024equiformerv2, qu2024importance, thölke2022torchmdnetequivarianttransformersneural}, it is important to note that \textit{they are not using the standard Transformer architecture} \citep{vaswani2023attentionneed} and still operate on a predefined graph structure. While a Transformer can be viewed as operating on a fully connected graph, its attention mechanism is fully connected for every input. In contrast, GNNs typically construct a new graph for each input (e.g., using a radius cutoff). Although some models have begun exploring Transformers for atomistic property prediction, they either use only textual molecular descriptors (e.g., SMILES) \citep{chithrananda2020chembertalargescaleselfsupervisedpretraining} or depend on custom graph embeddings \citep{kim2022puretransformerspowerfulgraph, rampášek2023recipegeneralpowerfulscalable} and modifications of the attention mechanism to include physical inductive biases \citep{zhou2023unimol, eissler2025simplegoofftheshelftransformer}. Broader ML and generative modeling work has also relaxed physical inductive biases \citep{ Abramson2024,joshi2025allatomdiffusiontransformersunified, wang2024swallowingbitterpillsimplified, vadgama2025probingequivariancesymmetrybreaking, flamshepherd2023languagemodelsgeneratemolecules}, but these approaches have not gained as much traction in molecular property prediction. In this study, we explore an unmodified Transformer without any graph structure or physics-inspired featurization.    

Frey et al. \citep{Frey2023} explored scaling in ML predictors of energies and forces but found significant deviations from consistent power law relationships with the models and datasets available at the time. More recently, Wood et al. \citep{wood2025umafamilyuniversalmodels} found scaling trends on the new OMol25 dataset \citep{levine2025openmolecules2025omol25}, but required a sophisticated mixture-of-experts scheme to train their largest models. To the best of our knowledge, there are no graph-based ML atomistic property predictors at the scale of models seen in other ML fields in terms of number of parameters \citep{touvron2023llamaopenefficientfoundation, brown2020languagemodelsfewshotlearners}.

Prior work has explored learning physical concepts from data in structural biology \citep{Roney2022} and with synthetic data \citep{lemos2022rediscoveringorbitalmechanicsmachine, de_Silva_2020}. We extend prior works to understand how the learned physics emerges as a function of data and model size using an architecture with no built in priors. 

\paragraph*{Graph neural networks (GNNs).} GNNs impose locality by applying computation on top of a pre-defined graph. The graph is constructed via a hard-coded algorithm, such as a radius cutoff. In contrast, Transformers always operate on a fully-connected graph for every input. While graph inductive bias aligns with the locality present in some atomistic prediction tasks, the impact of the graph topology in GNNs has been studied extensively \citep{rusch2023surveyoversmoothinggraphneural, digiovanni2023oversquashingmessagepassingneural, topping2022understandingoversquashingbottlenecksgraphs, dwivedi2023longrangegraphbenchmark, gat_oversmooth}, and evidence suggests that graphs can hurt performance and generalization Bechler-Speicher et al. \citep{bechlerspeicher2024gnnregular}. For instance, Kreiman and Krishnapriyan \citep{kreiman2025understandingmitigatingdistributionshifts} found that GNN-based atomistic property predictors tend to overfit to the graph structures encountered during training and struggle to generalize to new molecular geometries. These studies motivate us to investigate if the locality can be learned directly by a Transformer, without pre-defining a graph construction algorithm.

\section{Training an Unmodified Transformer without Physical Priors}
\label{sec:md_and_sl}

\begin{figure}
  \centering
  \includegraphics[width=0.6\linewidth]{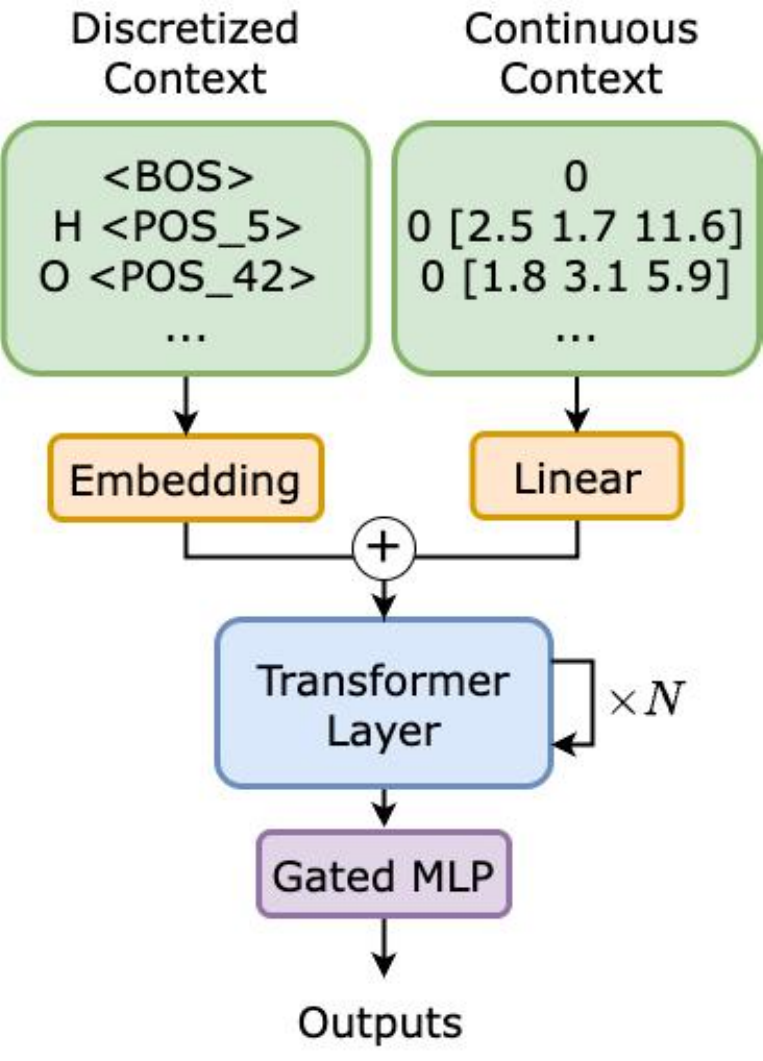}
  \caption{\textbf{A Graph-Free Transformer Design.} We investigate what physical principles can be learned by first training a neural network without any physical priors on atomistic data, and then probing its internal representations. Our model encodes both discretized and continuous molecular sequences using unmodified Transformer layers. Placeholder values represent discrete inputs in the continuous sequence. This contrasts with graph-based architectures that explicitly define an interaction graph between atoms using, for example, a radius cutoff.}
  \label{fig:model_fig}
\end{figure} 

We describe in detail how we train the Transformer model used in this work. We take ML molecular property prediction of energy and forces as a representative case study \citep{Unke2021} to study how ML models learn physics from data.  Traditionally, ML molecular property predictors incorporate physics-inspired features through graph-based message passing, where the molecular graph is constructed \textit{a priori} using heuristics such as a fixed radius cutoff or $k$-nearest neighbors. In this work, we remove these constraints entirely by replacing the GNN with a \textbf{standard Transformer} operating directly on Cartesian coordinates---without predefined graphs or chemistry-specific architectural modifications. This provides a clean test bed for studying whether physical priors and GNN-like relational patterns, can emerge naturally from data.

\paragraph*{Model architecture.} We use the LLaMA2 architecture \citep{touvron2023llama2openfoundation} as our backbone architecture, preserving the original multi-head self-attention mechanism. Our only architectural modifications are (1) removal of positional embeddings, since atomic positions are explicitly provided as input features and (2) the inclusion of both discrete and continuous embeddings for each token as model inputs, which are added before entering the attention layers. The attention mechanism itself remains \textit{completely unchanged}, ensuring that any learned relational structure arises from the data rather than from built-in inductive biases. \Cref{fig:model_fig} shows a schematic of the model architecture.  

\paragraph*{Input data format.}
\begin{figure*}
    \centering
    \includegraphics[width=0.9\linewidth]{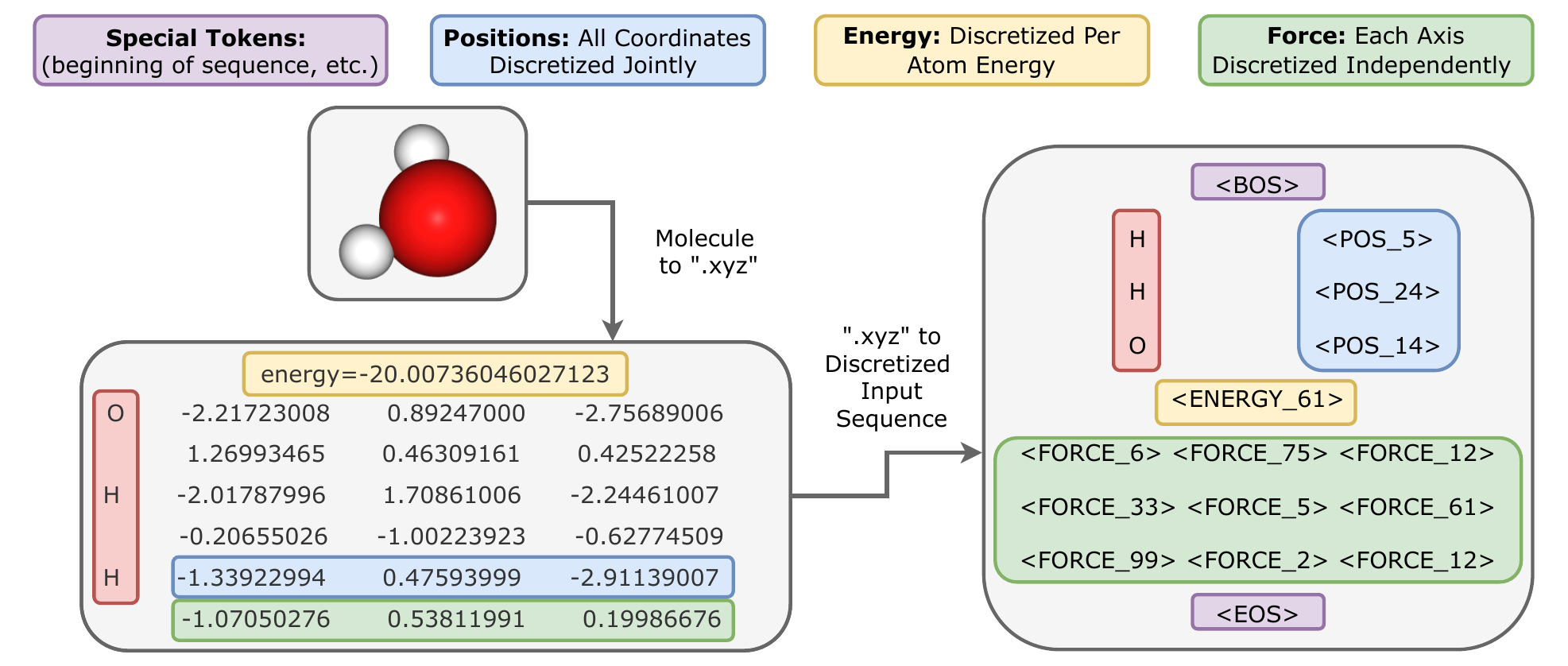}
    \caption{\textbf{Discretization Scheme during Pre-training.} We transform the standard ``.xyz'' molecular representation into a discretized input sequence for our model. To discretize continuous values, we use quantile binning, ensuring each bin contains the same number of datapoints. We jointly discretize atomic positions into a 3D grid, while force and energy components are discretized independently along each dimension. We add special tokens, like beginning and end of sequence tokens. Note that these discretized tokens are accompanied by the continuous values (for positions, forces, etc.), circumventing discretization errors for real-valued inputs.}
    \label{fig:disc_scheme}
\end{figure*}

Our model processes both a discretized and a continuous representation of each molecule’s \texttt{.xyz} file. Continuous molecular features such as positions, forces, and energies often exhibit heavy-tailed distributions spanning multiple orders of magnitude (\Cref{fig:pos_dist}), making using only a binned input prone to large discretization errors at the tails. Consequently, we give the model the continuous values on top of the discretized string, allowing it to rely on the continuous inputs for real-valued quantities.

As shown in \Cref{fig:disc_scheme}, we discretize continuous features using quantile binning so that each bin contains the same number of datapoints. The three position coordinates of each atom are jointly discretized into a three-dimensional grid, while force components are discretized independently along each axis. The continuous sequence provides the model with exact values for positions, forces, and energies, while using placeholder values to represent discrete inputs such as atomic numbers and special tokens. Special tokens mark the start and end of the sequence and indicate when the model should predict positions, forces, or energies. The embeddings for the continuous and discretized sequence are added before being input into the attention mechanism (see \Cref{apx:experiment_details} for more details). An example of a discretized input string is shown in \Cref{fig:disc_input}. 

\paragraph*{Training procedure} We adopt a two-stage training strategy inspired by pre‑train–then–fine‑tune approaches used by large language models~\citep{touvron2023llama2openfoundation, brown2020languagemodelsfewshotlearners, hoffmann2022trainingcomputeoptimallargelanguage}. During the pre-training stage, we train the model autoregressively with a causal attention mask to predict all discrete tokens in the sequence using a cross‑entropy loss. This allows it to learn the joint distribution of positions, forces, and energies, enabling likelihood estimation (\Cref{apx: further_exp}), and providing a strong initialization for downstream fine-tuning. 

During fine-tuning, the objective shifts to direct prediction of continuous energies and forces. We replace the causal attention mask with a bi-directional one, making the model permutation equivariant. We also replace the linear readout head that outputs a distribution over discrete tokens with two energy and force readout heads. The force head directly regresses the atomic embeddings to predict a force vector in $\R^{3}$ for each atom. The energy head predicts a per-atom energy which is aggregated across the system to get the total energy. During fine-tuning, the model operates in a continuous space, and no discretized tokens for energies or forces are included in the input.

We note that this training procedure, together with the results in the main text, provides an existence proof that Transformers can learn physical inductive biases directly from data. It is possible---if not likely---that future work will develop improved or simplified training strategies that yield even better accuracy, as has been observed across much of deep learning.

\section{Further Experiments}
\label{apx: further_exp}
We run further evaluations on OMol25 with our Transformer in \Cref{apx: md_transformer} before exploring the representations in more detail in \Cref{apx: more_reps}. Since during pre-training the Transformer learns the joint distribution over positions, forces, and energies, we then examine in \Cref{sec:uq} whether Transformers can be used for uncertainty quantification to identify where current ML atomistic property predictors make mistakes.  

\subsection{Further Evaluations on OMol25}
\label{apx: md_transformer}

\begin{figure}
    \centering
    \includegraphics[width=0.9\linewidth]{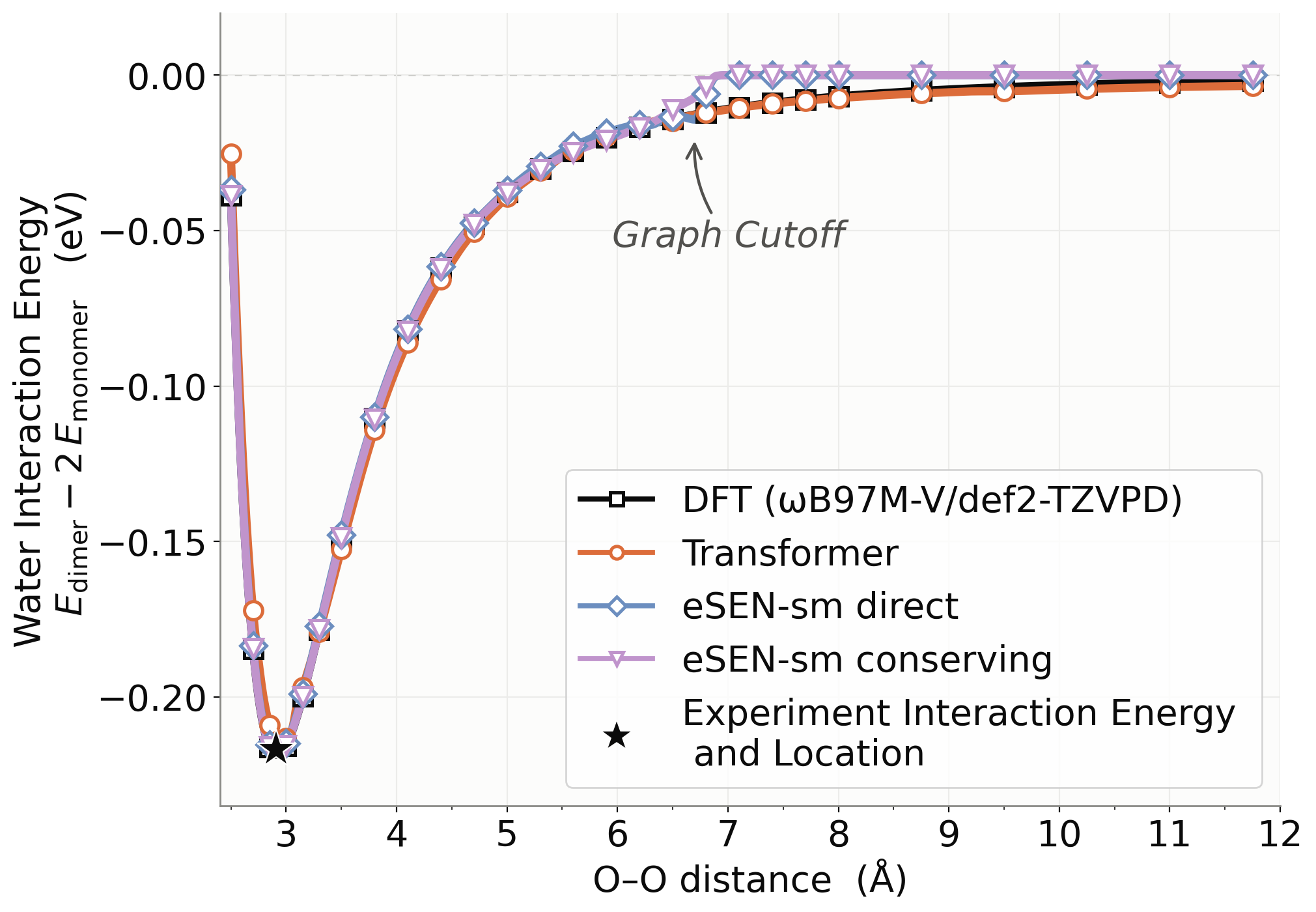}
    \caption{\update{\textbf{Transformers learn smooth interaction energies as a function of distance.} We plot the predicted water dimer interaction energy versus the oxygen-oxygen distance for the Transformer and eSEN-sm GNN, which incorporates physical priors such as a fixed-radius interaction cutoff, both trained on OMol25. We show the reference DFT curve, computed at the same level of theory as the OMol25 dataset, for comparison. All models accurately capture the interaction energy and the location of the energy minimum relative to DFT and experiment. The Transformer learns a smooth interaction energy without a premature jump caused by a fixed radius cutoff.}}
    \label{fig:water_dimer}
\end{figure}

\begin{figure*}
    \centering
    \includegraphics[width=0.9\linewidth]{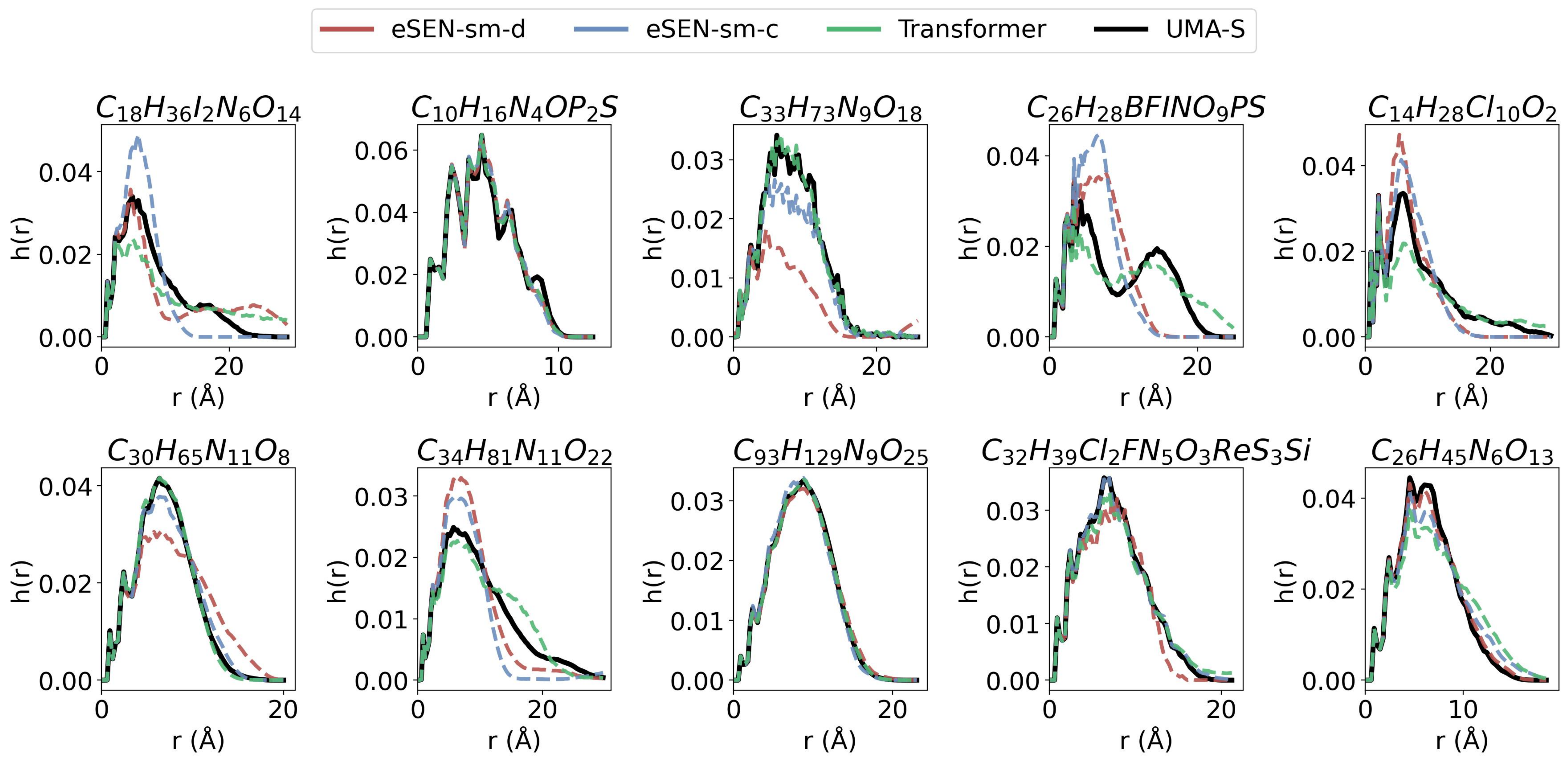}
    \caption{\textbf{Transformers can accurately run MD simulations relative to models with strong inductive biases.} We select 10 random molecules from the OMol validation set and run NVT MD simulations at 500K for 100ps using a 0.5 fs timestep. We plot the estimated distribution of interatomic distances $h(r)$ for the eSEN-sm models (both direct and conserving versions) and the UMA-s model as reference. Transformers accurately reproduce the distribution of interatomic distances relative to these models without using any physical inductive biases.}
    \label{fig:hr_omol}
\end{figure*}

Beyond learning consistent physical patterns, we examine how well the model is able to do on the original quantum mechanical energy and force prediction task. We train a 1B parameter Transformer on the OMol25 4 million (4M) training split, using the same total training compute budget (measured in FLOPs) as eSEN---a state-of-the-art 6M-parameter equivariant GNN  model which serves as our point of reference.\footnote{FLOPs for eSEN are estimated using the FairChem repository; FLOPs for our Transformer are estimated via HuggingFace tooling.}

We compare the energy and force mean absolute errors (MAE) for our Transformer to those of eSEN, as well as the training and inference speeds, in \Cref{tab:omol_val_total}. Despite having no built-in geometric priors, the Transformer achieves competitive accuracy with eSEN on energies and forces. It also trains and runs faster in wall‑clock time, benefiting from mature software and hardware optimizations for Transformer architectures~\citep{pytorch2, jouppi2017tpu}. We emphasize that both models are trained on the same data and that OMol25 consists of the most diverse quantum mechanical validation set to date, covering biomolecules, electrolytes, metal complexes, and beyond.

\begin{table*}[h!]
\caption{\textbf{Out-of-distribution composition validation results.} Transformers match the energy and force errors of a state-of-the-art equivariant GNN \citep{fu2025learningsmoothexpressiveinteratomic_esen, levine2025openmolecules2025omol25} under the same training computational budget, while achieving faster training and inference in wall clock time. We estimate FLOPs through the FairChem repository for eSEN and through HuggingFace tooling for our Transformer. We measure the training speed on a single node of 4 H100s and the forward latency on a single A6000 with a system of 100 atoms.}
\centering
\resizebox{\textwidth}{!}{%
\begin{tabular}{l c c c cc}
\toprule
Model & FLOPs & Forward Latency (ms) & Training Speed  (atoms/sec) & Energy MAE (meV) & Forces MAE (meV/Å) \\
\midrule
eSEN-sm-d 6M      & $O(10^{20})$             & 26.3 & 32k+ & 129.77 & 13.01 \\
\cmidrule(l){1-6}
Transformer 1B (Ours)      & $8.5\times10^{19}$       & 17.2 & 42k+ & 117.99     & 18.35    \\
\bottomrule
\end{tabular}%
}
\label{tab:omol_val_total}
\vspace{-6pt}
\end{table*}

\paragraph*{Why is the Transformer faster despite having so many more parameters?} 
There are many ways to compare the efficiency of models. Comparisons can be made between model parameters, model FLOPs, or wall-clock time. These can each be misleading for their own reasons, and we provided measures of each in \Cref{tab:omol_val_total}. Raw parameters can be misleading since parameters can be used multiple times during a model's forward pass. For example, GNNs often use parameters multiple times per node and edge to construct messages in the message passing step. FLOPs alone can be misleading when comparing different model types since different types of operations can be implemented at different speeds on modern hardware. For example, a sequential operation could be slower compared to a parallel one even if they have the same number of FLOPs, and sparse operations (like those in GNNs) are often slower to implement than dense ones (like those in Transformers). Finally, wall-clock time is system dependent and can improve with the next generation of hardware. Regardless, we find that Transformers leverage mature software and hardware frameworks to run efficiently, even compared to GNNs with far fewer parameters. Although the asymptotic complexity is different between the Transformer and GNN, the improved efficiency of modern hardware paired with the fact that large systems can be spliced into smaller chunks means that the complexity does not cause issues for many relevant problem scales.    

We also evaluate whether Transformers can learn rotational equivariance from data alone by measuring the cosine similarity between forces predicted in different rotational frames: $
    \textnormal{cossim}(\mathbf{R}\mathbf{F}(\mathbf{r}), \mathbf{F}(\mathbf{R}\mathbf{r})),
$ where $R$ is a rotation matrix and $\mathbf{F}(\vr)$ are the model's predicted forces for a system with atomic positions $\vr$. Averaged over the OMol25 validation set, the similarity exceeds $0.99$, consistent with prior findings~\citep{qu2024importance, neumann2024orbfastscalableneural, eissler2025simplegoofftheshelftransformer} that models without explicit equivariance can learn approximate equivariance directly from training data. As an extreme test of symmetry learning, we evaluate whether the model can recover permutation invariance when atom ordering information is explicitly introduced. To do so, we inject positional embeddings \citep{su2023roformerenhancedtransformerrotary}, thereby imposing an artificial ordering over atoms. Surprisingly, despite this symmetry-breaking modification, the average total energy MAE between randomly permuted atomic sequences is less than $5$ meV on the OMol25 validation set---corresponding to less than $5\%$ of the model’s error relative to the reference DFT labels. This result indicates that the model can also learn this permutation invariance from data.\footnote{We are not suggesting that permutation invariance should be omitted in practice, as it is straightforward to impose architecturally. Rather, we emphasize that its emergence under these conditions highlights the capacity of data-driven training to recover fundamental physical symmetries.} We note that these values can essentially arbitrarily improve through frame averaging \citep{puny2022frameaveraginginvariantequivariant, duval2023faenetframeaveragingequivariant}, though at the cost of slower inference.

To further evaluate the Transformer trained on OMol25, we run MD simulations using ten randomly sampled molecules from the OMol25 validation set. We first run NVT simulations and calculate a thermodynamic observable (the distribution of interatomic distances) from the simulations to evaluate the quality of the dynamics. The distribution of interatomic distances is a commonly used observable which characterizes 3D molecular structures \citep{Zhang_2018, fu2023forces, raja2025stabilityaware} and is defined as:
\begin{equation}
    \label{eqn:hr}
    h(r) = \frac{1}{n(n-1)} \sum_{i=1}^n \sum_{i \neq j} \delta(r - ||\mathbf{X}_i - \mathbf{X}_j||),
\end{equation}
where $\mathbf{X} \in \R^{n\times3}$ are the positions of the $n$ atoms of a molecule and $\delta$ is the Dirac Delta function. For our evaluation metric, we calculate the $h(r)$ MAE with respect to a reference:
\begin{equation}
    \label{eqn:hr_mae}
    \int_{r=0}^{\infty} dr |\langle h^{*}(r) \rangle - \langle \hat{h}(r) \rangle|,
\end{equation}
where $\langle \cdot \rangle$ represents an average over structures sampled from the predicted ($\hat{h}(r)$) or reference ($h^{*}(r)$) Boltzmann distribution.

We simulate 10 random validation molecules for $100$ps using a 0.5 fs timestep. Following previous work, we use a Langevin thermostat at 500K with a friction of 0.01 fs$^{-1}$ \citep{fu2023forces, kreiman2025understandingmitigatingdistributionshifts}. We also run simulations with the eSEN-sm models (both conserving and direct prediction versions) \citep{levine2025openmolecules2025omol25}. We use simulations from the UMA-S model \citep{wood2025umafamilyuniversalmodels} as a reference since it is a larger model trained on significantly more data. The Transformer has a $h(r)$ MAE (see \Cref{eqn:hr_mae}) of 0.040 relative to UMA-S for the 10 molecules (compared to 0.077 and 0.065 for eSEN-sm-d and eSEN-sm-c, respectively), showing that Transformers can be applied to run molecular dynamics (see \Cref{fig:hr_omol}).

\begin{figure}
    \centering
    \includegraphics[width=\linewidth]{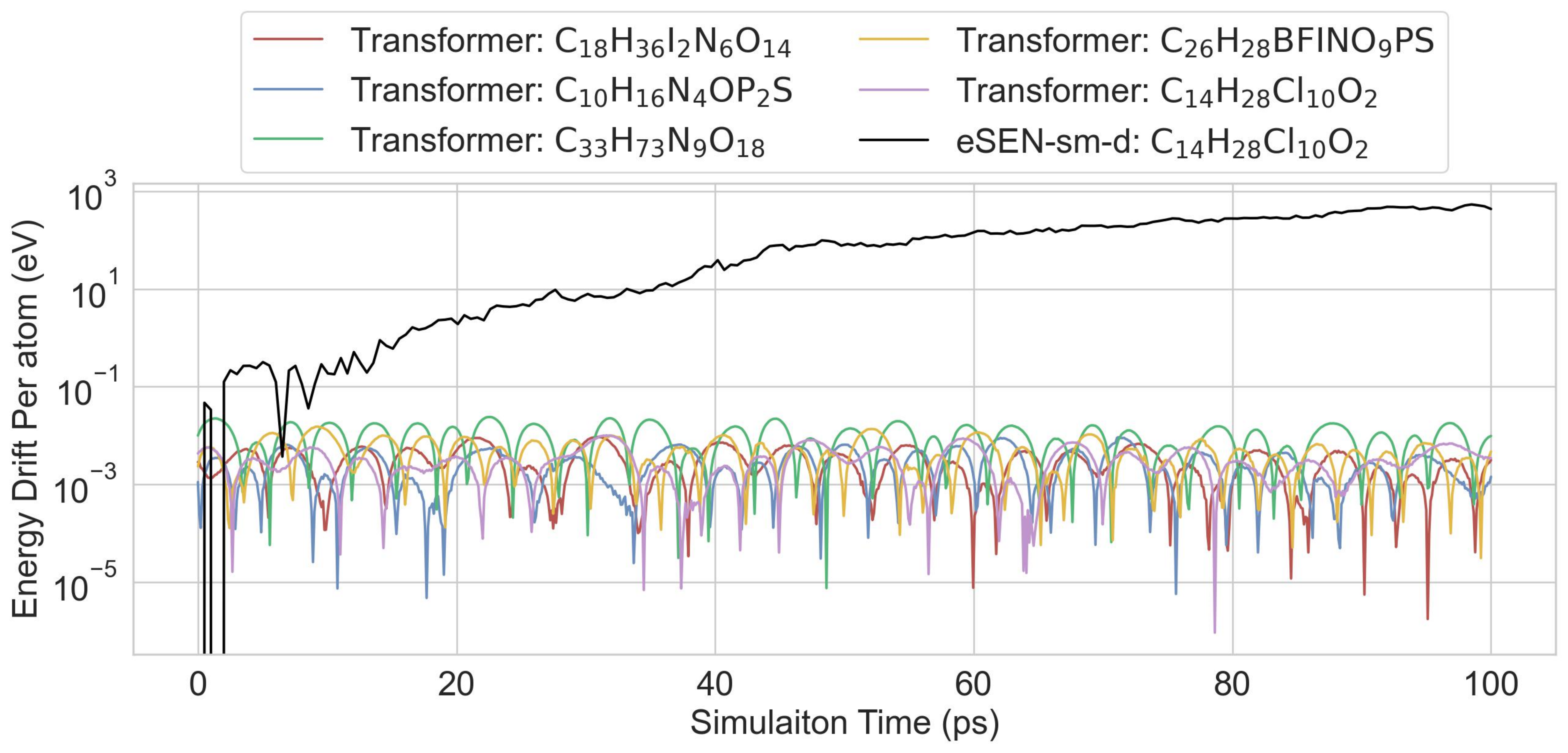}
    \caption{\textbf{Transformers can conserve energy during NVE simulations when fine-tuned to predict conservative forces.}}
    \label{fig:nve_sim}
\end{figure}

Since conservative force fields are important for many downstream applications, we also explore fine-tuning our Transformer to predict forces as the gradient of the energy. We selected the first 5 molecules used to evaluate the NVT simulations and additionally ran NVE simulations for $100$ps using a $0.5$fs timestep. The fine-tuned Transformer was able to accurately conserve energy, whereas a direct-prediction model experiences significant energy drift (see \Cref{fig:nve_sim}). While we observed that fine-tuning with energy gradients made training more unstable, this provides a proof of concept that graph-free Transformers can still be accurately used for downstream molecular tasks. We think it is an interesting direction for future work to examine how other physical constraints can be taught to Transformers after the initial training phase. While more rigorous evaluation would be needed before deployment in large-scale scientific discovery, these results demonstrate that a graph-free Transformer can make physically meaningful predictions---making it a compelling case study for the representation-learning analysis done in the main text.

\update{Beyond looking at interatomic interaction patterns, we also analyze a simple water dimer system to evaluate learned \emph{intermolecular} interactions. In \Cref{fig:water_dimer}, we see that both the Transformer and eSEN-sm models, both trained on OMol25, accurately predict the interaction energy curve \citep{RocherCasterline2011}. The Transformer implicitly learns a smooth decay in the interaction energy, without succumbing to an abrupt truncation of interactions caused by a fixed radius cutoff.}

\subsection{Analysis of Learned Representations of Graph-Free Transformers}
\label{apx: more_reps}

\begin{figure}
    \centering
    \includegraphics[width=\linewidth]{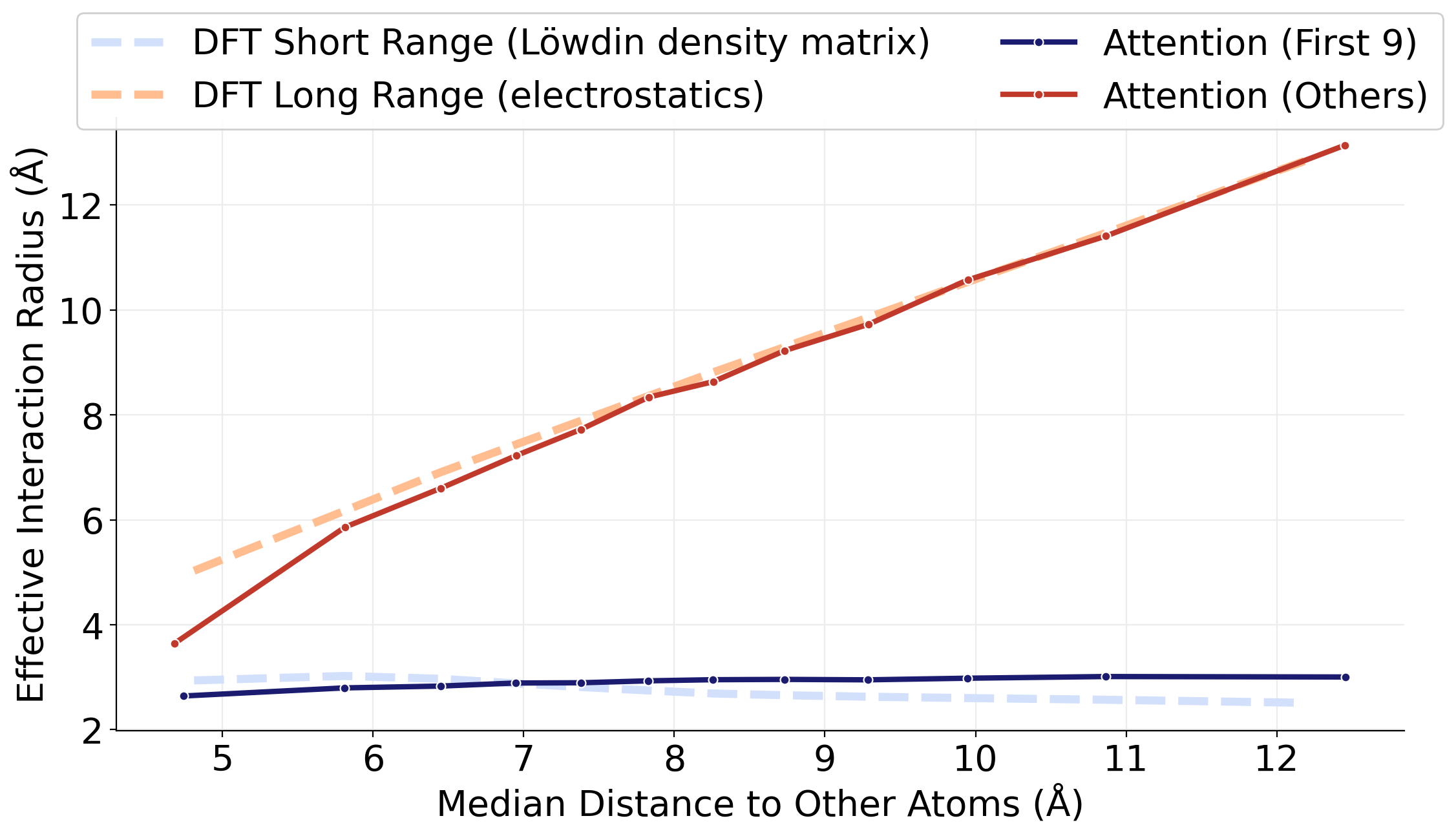}
    \caption{\update{\textbf{Transformers adapt their effective receptive field based on atomic environment.} For each atom, we define the model's learned effective interaction radius as the smallest distance containing 90\% of its attention mass. We measure local atomic environment by the median distance to all other atoms. We compute DFT-derived references using the same 90\% radius definition for two pairwise proxies: a short-range electronic-delocalization proxy from the L\"owdin-orthogonalized density matrix and a long-range monopole electrostatic proxy from L\"owdin partial charges. Early layers align with the short-range reference with an effective interaction radius roughly constant across atomic environments. Later layers learn an effective interaction radius that increases as an atom becomes more isolated, aligning with the long-range electrostatic reference.}} 
    \label{fig:attn_effective_radius}
\end{figure}%

\begin{figure}
    \centering
    \includegraphics[width=0.8\linewidth]{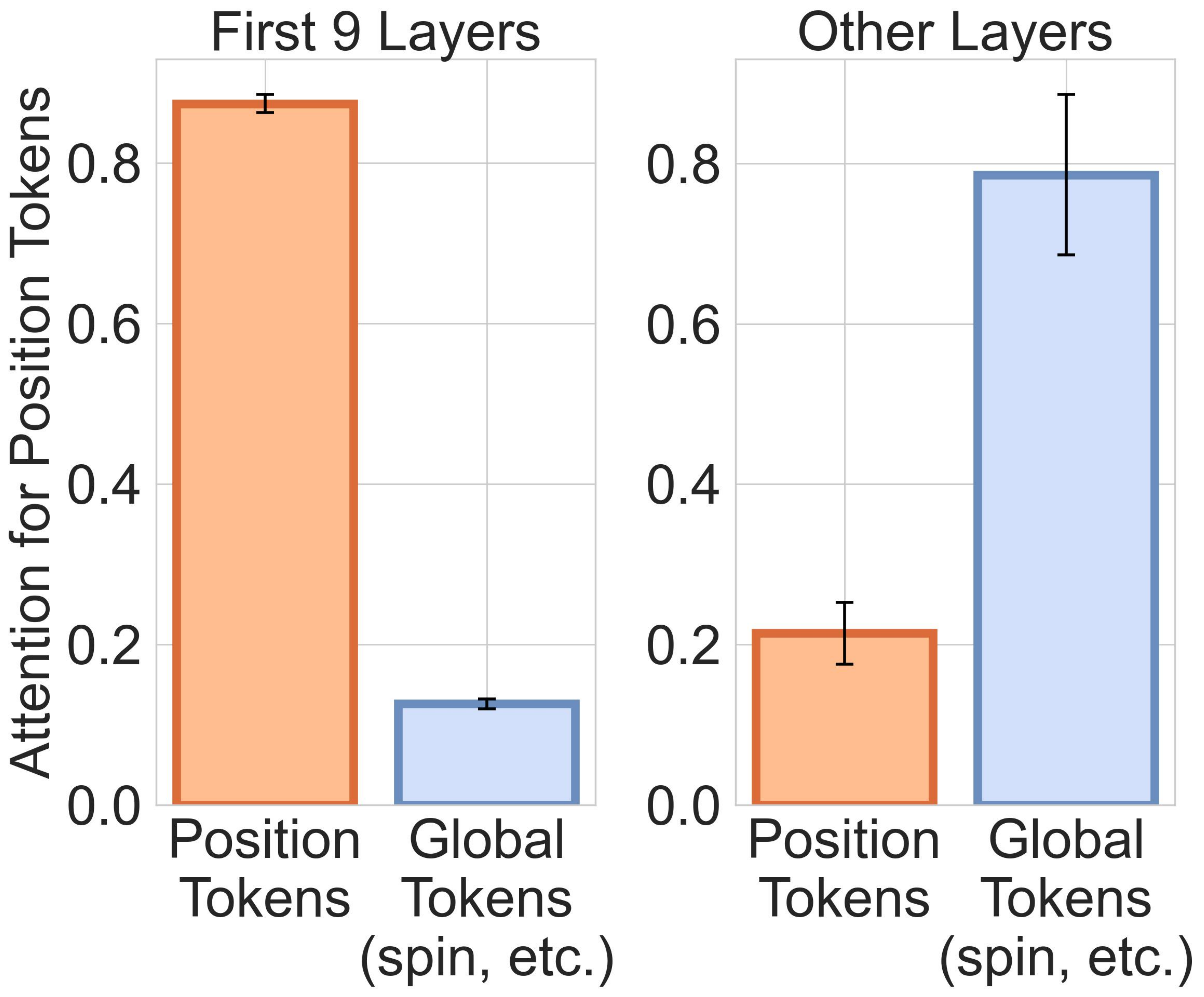}
    \caption{\update{\textbf{\textbf{Attention Distribution by Token Type.}} We show what fraction of attention from position tokens goes towards other position tokens versus global tokens, such as charge and spin. In the first nine layers, position tokens predominantly focus on other position tokens. In the later layers, attention shifts towards global tokens. Paired with the electrostatic-like relationship between attention score and interatomic distance,  these results suggest that Transformers capture local features in the early layers and then aggregate global information in the final layers.}}
    \label{fig:attn_by_type}
\end{figure}

\begin{figure*}
    \centering
    \includegraphics[width=\linewidth]{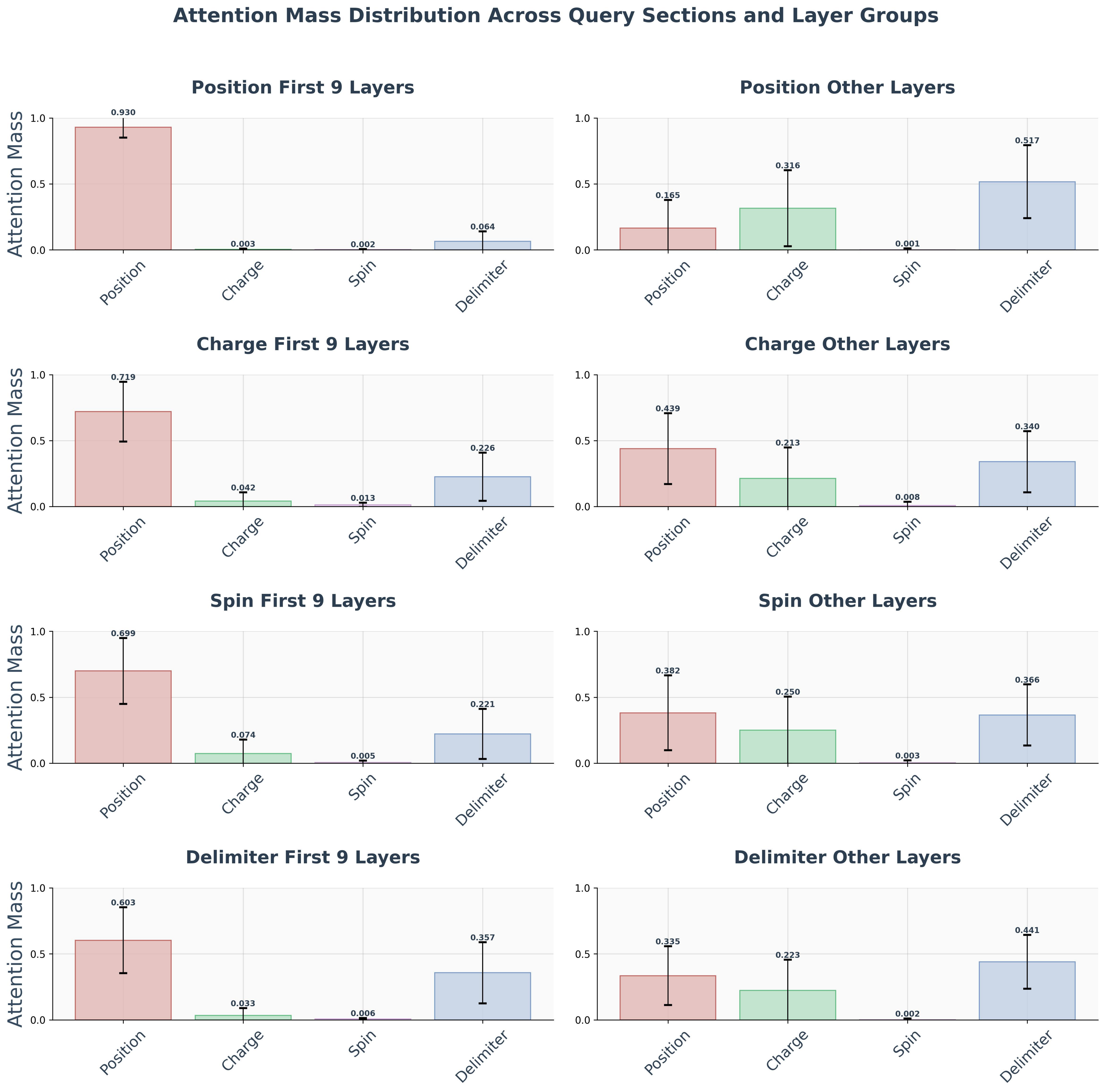}
    \caption{\textbf{Interatomic attention dynamics dominate in early layers while global attention dominates in later layers.} In the first $9$ layers, position tokens attend almost exclusively to the other position tokens. They then shift attention to other input tokens in later layers, suggesting that they're accessing global graph information. We observe that barely any attention is paid to the spin tokens. We attribute this to the fact that high-spin molecules are rare in the training dataset (see \Cref{fig:pos_dist}), suggesting that the model needs more varied spin data to learn how to use spin effectively.}
    \label{fig:attn_mass_combined_all_sections}
\end{figure*}

We probe the internal representations of our Transformer to understand what physical structure it has learned from training to predict energies and forces on OMol25. Specifically, we provide more detailed plots for our attention score analysis. We breakdown by layer in \Cref{fig:attn_vs_dist_per_layer} the relationship between attention score and interatomic distance. We also provide a breakdown of the attention distribution by token type in \Cref{fig:attn_by_type}, where we see that attention focuses on positions in early layers and switches to global tokens in later layers. In \Cref{fig:attn_mass_combined_all_sections}, we further subdivide tokens and group the inputs into four semantic buckets of \textbf{Positions}, \textbf{Charge}, \textbf{Spin}, and \textbf{Delimiter} tokens. The first three buckets encode the atomic positions, the molecular charge, and the molecular spin, respectively. The last bucket includes all other tokens, which encode delimiter information of where each section begins and ends (e.g., $\text{[POS], [POS\_END]}$ tokens). We then produce four figures, each restricting rows of the attention matrix to each of the four buckets, respectively. For each figure, we plot the evolution of the attention mass distribution across each of the four buckets, as we move from the first $9$ layers to later layers (see \Cref{fig:attn_mass_combined_all_sections}). In the first $9$ layers, \textbf{Positions} tokens attend almost exclusively to the other \textbf{Positions} tokens. They then shift attention to other input tokens in later layers. This is consistent with the picture observed earlier that in layers $1$-$9$ interatomic attention dynamics dominate, while it dies down in later layers. It is natural to ask what relevant information these non-position tokens carry, in addition to charge and spin information, by the time \textbf{Positions} tokens shift attention to them. To this end, if we take a look at the mass distribution of \textbf{Charge}, \textbf{Spin}, and \textbf{Delimiter} tokens (last three rows of \Cref{fig:attn_mass_combined_all_sections}), we notice that all of them predominantly ($ > 60\%$) attend to \textbf{Positions} tokens in the first $9$ layers. This suggests non-position tokens aggregate global graph information in layers $1$-$9$, acting as information banks. \textbf{Positions} tokens, after they shift focus from interatomic attention, then access this global information in later layers by attending to them. We acknowledge that it appears that virtually no attention is given to \textbf{Spin} tokens. We attribute this to the fact that high-spin molecules are exceedingly rare in the training dataset (see \Cref{fig:pos_dist}), suggesting that the model needs more exposure to learn how to use spin effectively.

\update{\paragraph*{Learned adaptive interaction radius.} We investigate whether the Transformers learned interaction strength is dependent on atomic environment. For each atom in the OMol25 validation set, we define the model's learned effective interaction radius as the smallest distance from atom $i$ that contains 90\% of its total attention mass (see Eqn. 2 in the main text or \Cref{eqn:effective_rad_redefined} below). In \Cref{fig:attn_effective_radius} and \Cref{fig:attn_radius_vs_med_distance_by_layer}, we compare the model's learned interaction scale---represented by the attention scores---to a measure of local sparsity, defined as the median distance from $i$ to all other atoms in the structure. The Transformer learns to increase its effective receptive field for more isolated atoms.}

\update{For reference, we compute two analogous DFT-derived interaction radii from the raw ORCA outputs in the OMol25 dataset \citep{levine2025openmolecules2025omol25}. We compute a short-range reference based on the L\"owdin-orthogonalized one-particle density matrix. Let $P$ denote the one-particle density matrix in the atomic-orbital basis and $S$ the atomic-orbital overlap matrix:}
  $$\widetilde{P} = S^{1/2} P S^{1/2},$$
  $$w^{\mathrm{sr}}_{ij}
  =
  \sum_{\mu \in i,\, \nu \in j}
  \left|
  \widetilde{P}_{\mu\nu}
  \right|,
  $$
\update{where $\mu$ and $\nu$ index atomic orbitals centered on atoms $i$ and $j$, respectively. This quantity measures electronic delocalization or quantum-mechanical coupling between atoms, and provides a nearsighted short-range reference \citep{Wiberg1968ApplicationOT, Mayer1983ChargeBO, Prodan2005}. For a long-range reference, we use a monopole electrostatic proxy computed from L\"owdin partial charges. For atoms $i$ and $j$ with charges $q_i$ and $q_j$ separated by distance $r_{ij}$, we define}
  $$
  w^{\mathrm{lr}}_{ij}
  =
  \frac{|q_i q_j|}{r_{ij}}.
  $$
\update{The absolute value removes sign cancellations so that the resulting radius measures the spatial extent of electrostatic interaction magnitude rather than the net electrostatic energy. For either DFT-derived proxy, $w_{ij}$ denotes the pair strength between atom $i$ and atom $j$. The corresponding 90\% reference radius for atom $i$ is obtained by sorting all atoms by increasing distance $r_{ij}$ and finding the smallest radius $R_i$ such that}
\begin{equation}
\label{eqn:effective_rad_redefined}
  \sum_{r_{ij} \leq R_i} w_{ij}
  \geq
  0.9
  \sum w_{ij}.  
\end{equation}

\update{This gives a common effective-radius definition for model attention, density-matrix coupling, and electrostatic coupling. \Cref{fig:attn_effective_radius} shows how early model layers have an effective radius that is nearly independent of neighborhood sparsity and close to the short-range density-matrix reference. Later layers show an increasing radius for more isolated atoms and track the long-range electrostatic reference. This suggests that the model can adapt its effective interaction radius per atom based on the local atomic environment, and provides further insights into how the model handles both long- and short-range phenomena.}

\begin{figure*}
    \centering
    \includegraphics[width=0.85\linewidth]{images/attn_vs_dist_per_layer.pdf}
    \caption{\textbf{Attention is strongly inversely correlated with distance in layers $1$-$9$.} Attention decays steeply with increasing interatomic distance in layers $1$-$9$. This implies that the model is learning to attend predominantly to local interactions in early layers.}
    \label{fig:attn_vs_dist_per_layer}
\end{figure*}

\begin{figure*}
    \centering
    \includegraphics[width=\linewidth]{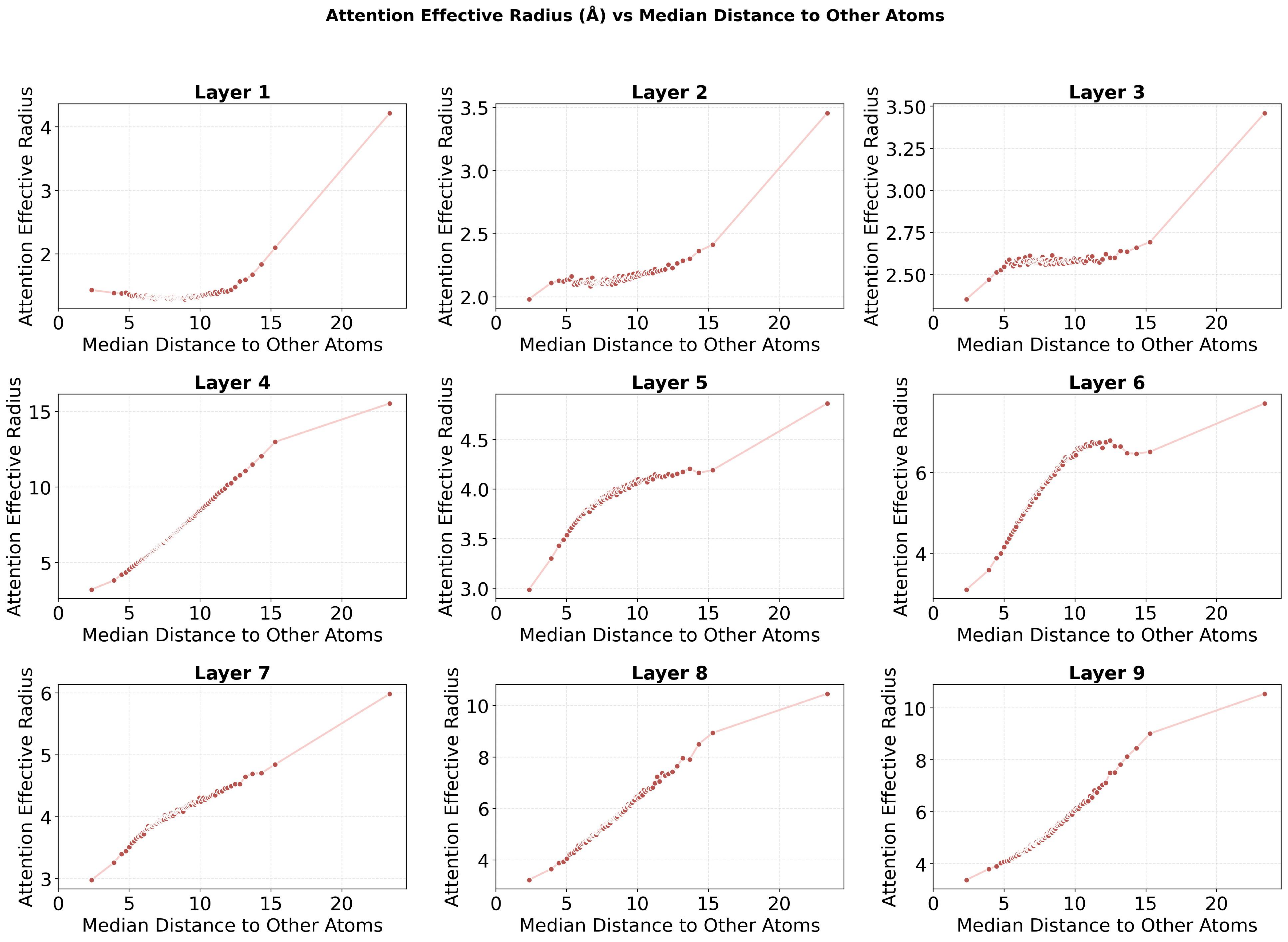}
    \caption{\textbf{Effective attention radius is adaptive to how tightly packed an atom is within a molecule.} We observe a clear positive trend between effective attention radius and median distance to neighbors. Within certain layers (e.g., layer $4$), radius can go as high as $15$Å and as low as $2$Å depending on the molecule and atom within that molecule.}
    \label{fig:attn_radius_vs_med_distance_by_layer}
\end{figure*}

\begin{figure*}[t!]
    \centering
    
    \begin{subfigure}[t]{0.46\textwidth}
        \centering
        \includegraphics[width=\linewidth]{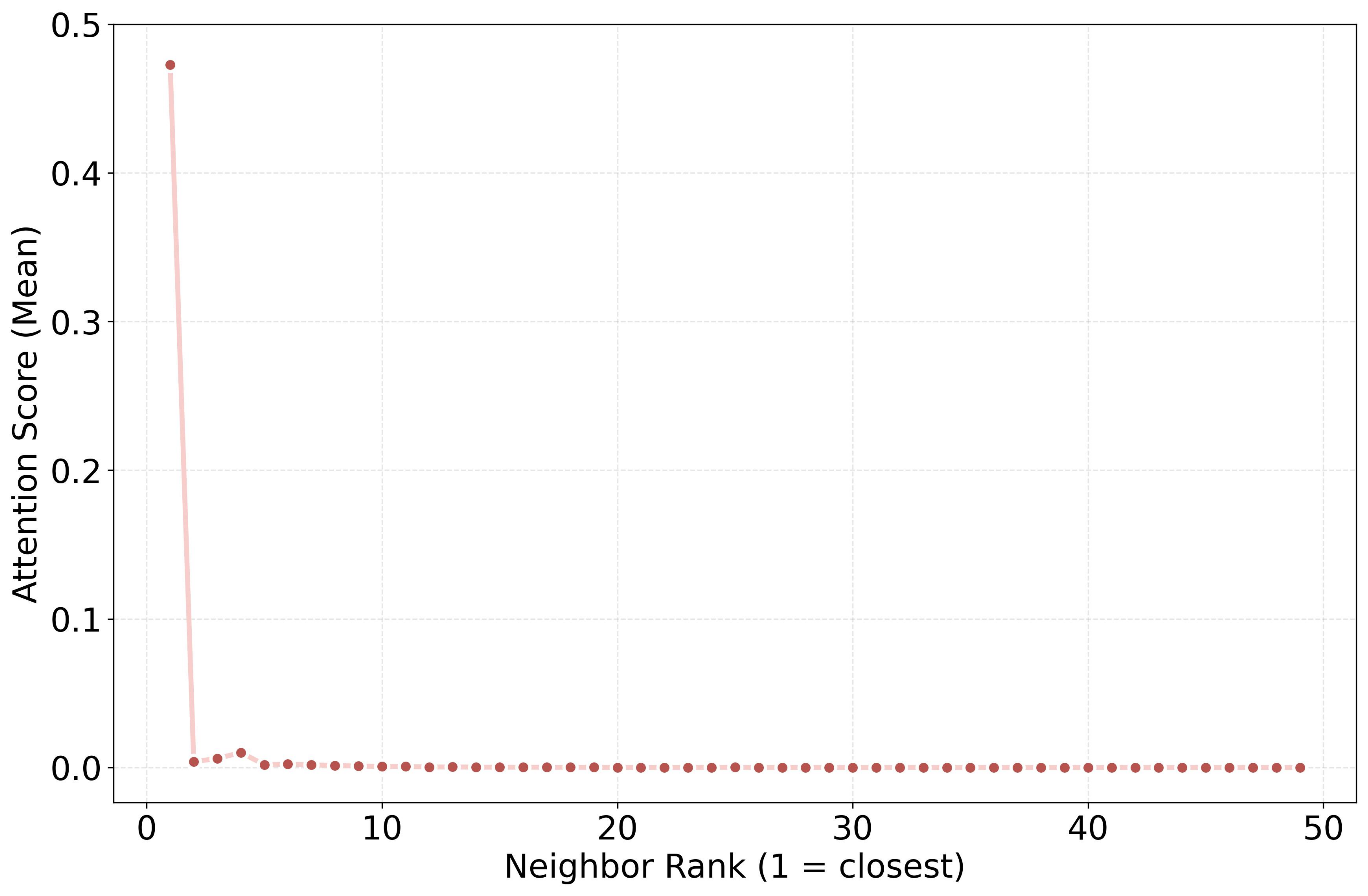}
        \caption{\textbf{Layer 1 Head 1}}
        \label{fig:attn_vs_rank_layer_1_head_1}
    \end{subfigure}%
    \hspace{6pt}
    \begin{subfigure}[t]{0.46\textwidth}
        \centering
        \includegraphics[width=\linewidth]{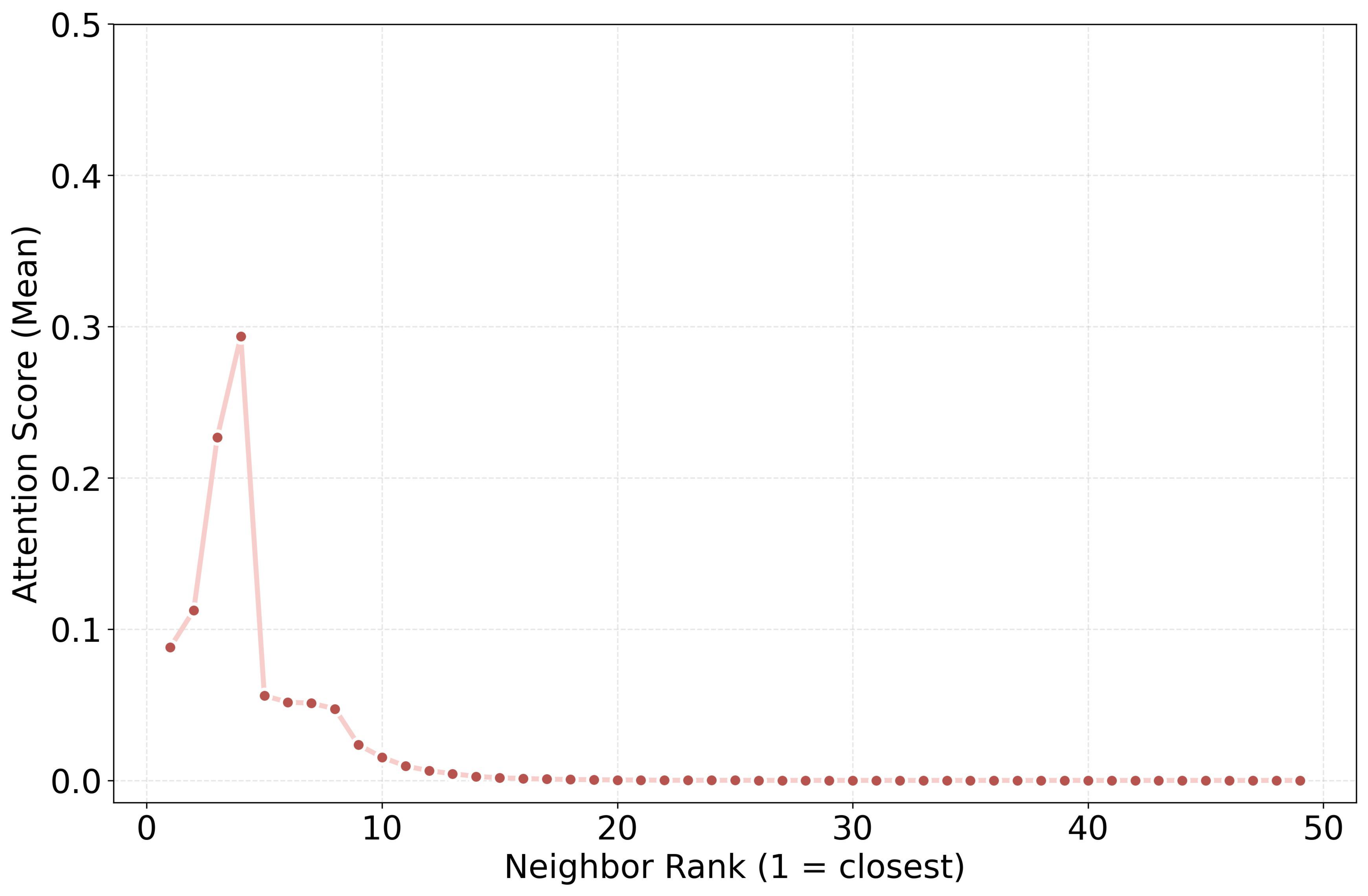}
        \caption{\textbf{Layer 8 Head 0}}
        \label{fig:attn_vs_rank_layer_8_head_0}
    \end{subfigure}%
    \caption{\textbf{Certain heads in certain layers exhibit unique rank-based attention behavior.} (a)  Layer $1$ head $1$ exhibits a $1$-nearest-neighbor pattern where attention is overwhelmingly placed on the closest neighbor and almost no attention beyond. (b) Layer $8$ head $0$ increases attention with rank and peaks at rank 4, before it drops sharply, and decreases gradually after.}
    \label{fig: special_knn_behavior}
\end{figure*}

\begin{figure*}[t!]
    \centering
    
    \begin{subfigure}[t]{0.46\textwidth}
        \centering
        \includegraphics[width=\linewidth]{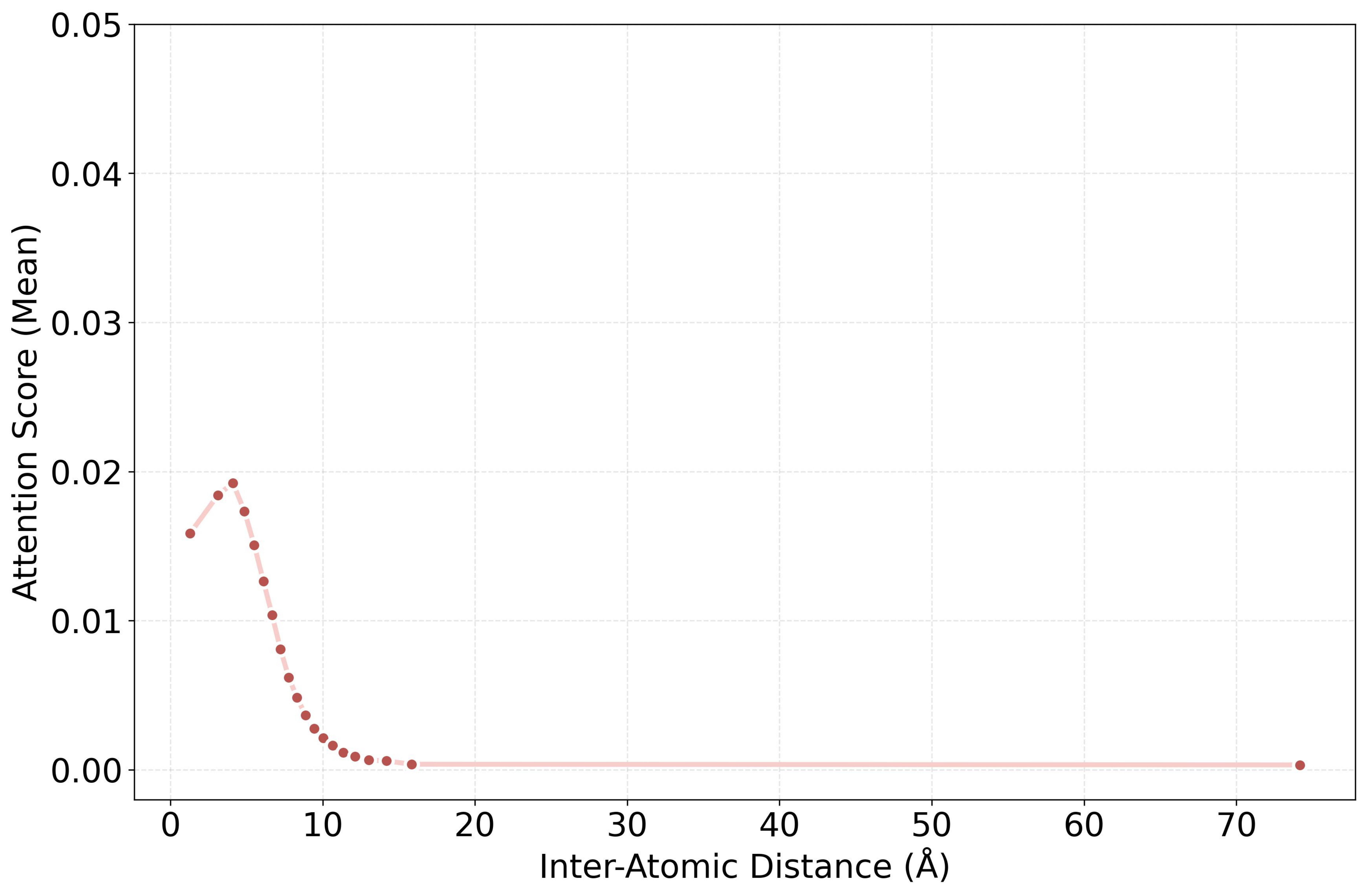}
        \caption{\textbf{Layer 8 Head 1}}
        \label{fig:attn_vs_dist_layer_8_head_1}
    \end{subfigure}%
    \hspace{6pt}
    \begin{subfigure}[t]{0.46\textwidth}
        \centering
        \includegraphics[width=\linewidth]{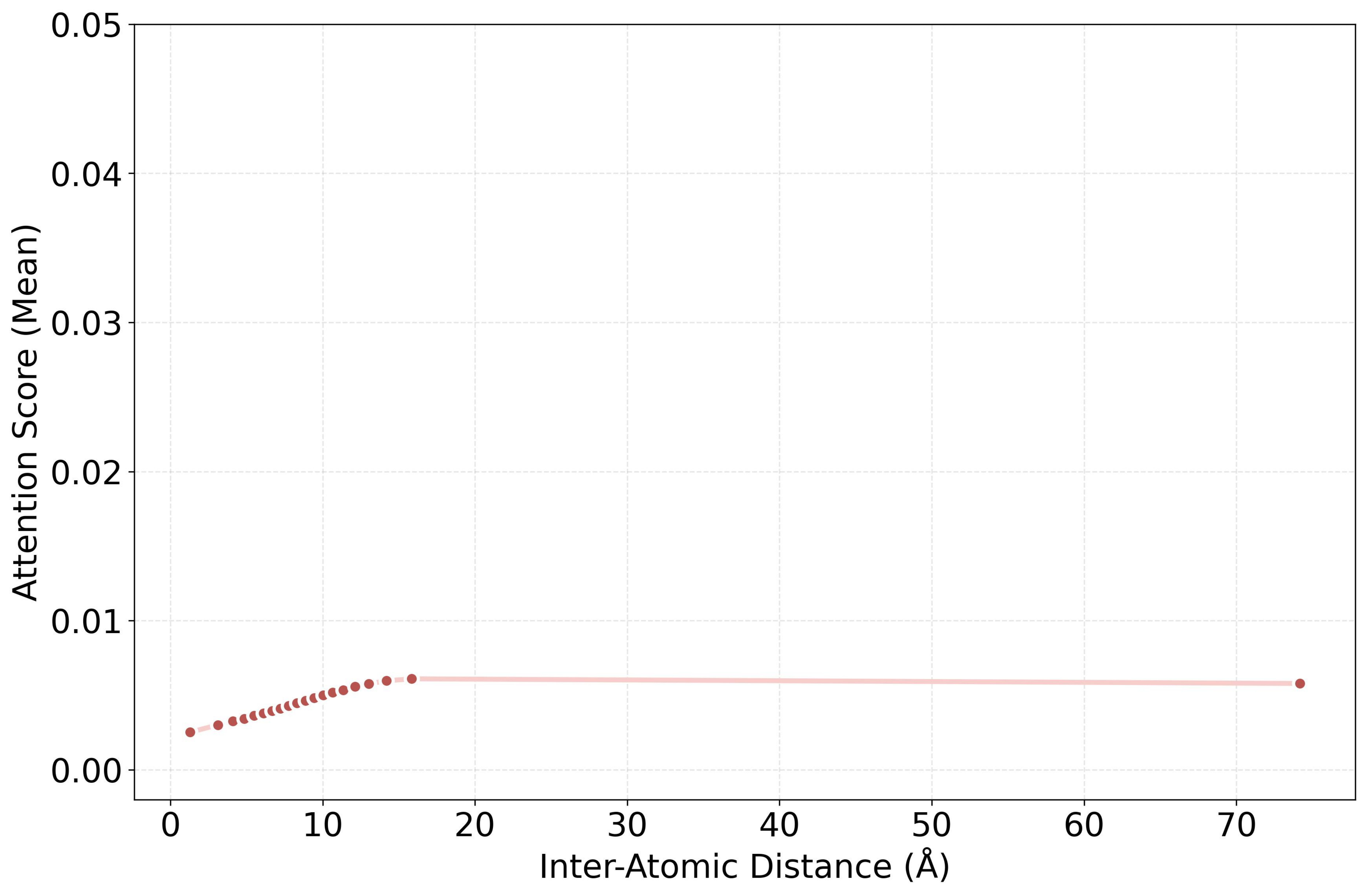}
        \caption{\textbf{Layer 6 Head 6}}
        \label{fig:attn_vs_dist_layer_6_head_6}
    \end{subfigure}%
    \caption{\textbf{Certain heads in certain layers exhibit unique attention relationships with distance.} (a)  Layer $8$ head $1$ exhibits non-monotonic attention that increases at small distances, peaks at $\sim 5$Å, then decays gradually. (b) Layer $6$ head $6$ exhibits positive correlation of attention with distance, where attention increases steadily putting more mass on far-away atoms.}
    \label{fig: special_dist_behavior}
\end{figure*}

\paragraph*{Unique attention patterns.} Finally, we end this section by observing that certain heads in certain layers exhibit unique interatomic attention behavior that deviates from the smooth monotonic decay of attention with distance we observe when we average over heads. Such patterns are evident when we plot against neighbor absolute distance, as well as rank. We note two examples that stand out in each domain. In \Cref{fig:attn_vs_rank_layer_1_head_1}, layer $1$ head $1$ exhibits a $1$-nearest-neighbor pattern where attention is overwhelmingly placed on the closest neighbor and almost no attention beyond. And in \Cref{fig:attn_vs_rank_layer_8_head_0}, layer $8$ head $0$ increases attention with rank, peaks at rank 4, before it drops sharply, and 
decreases gradually after.  In \Cref{fig:attn_vs_dist_layer_8_head_1}, layer $8$ head $1$ exhibits non-monotonic attention that increases at small distances, peaks at $\sim 5$Å, then decays gradually. In \Cref{fig:attn_vs_dist_layer_6_head_6}, Layer $6$ head $6$ exhibits positive correlation of attention with distance, where attention increases steadily, putting more mass on distant atoms. These examples demonstrate non-monotonic and long-range interatomic attention patterns, which appear to be useful for molecular energy and force prediction, but are difficult to anticipate or hard-code \textit{a priori}.

\subsection{Pre-training Scaling Laws}

\begin{figure}[H]
    \centering
    \includegraphics[width=0.9\linewidth]{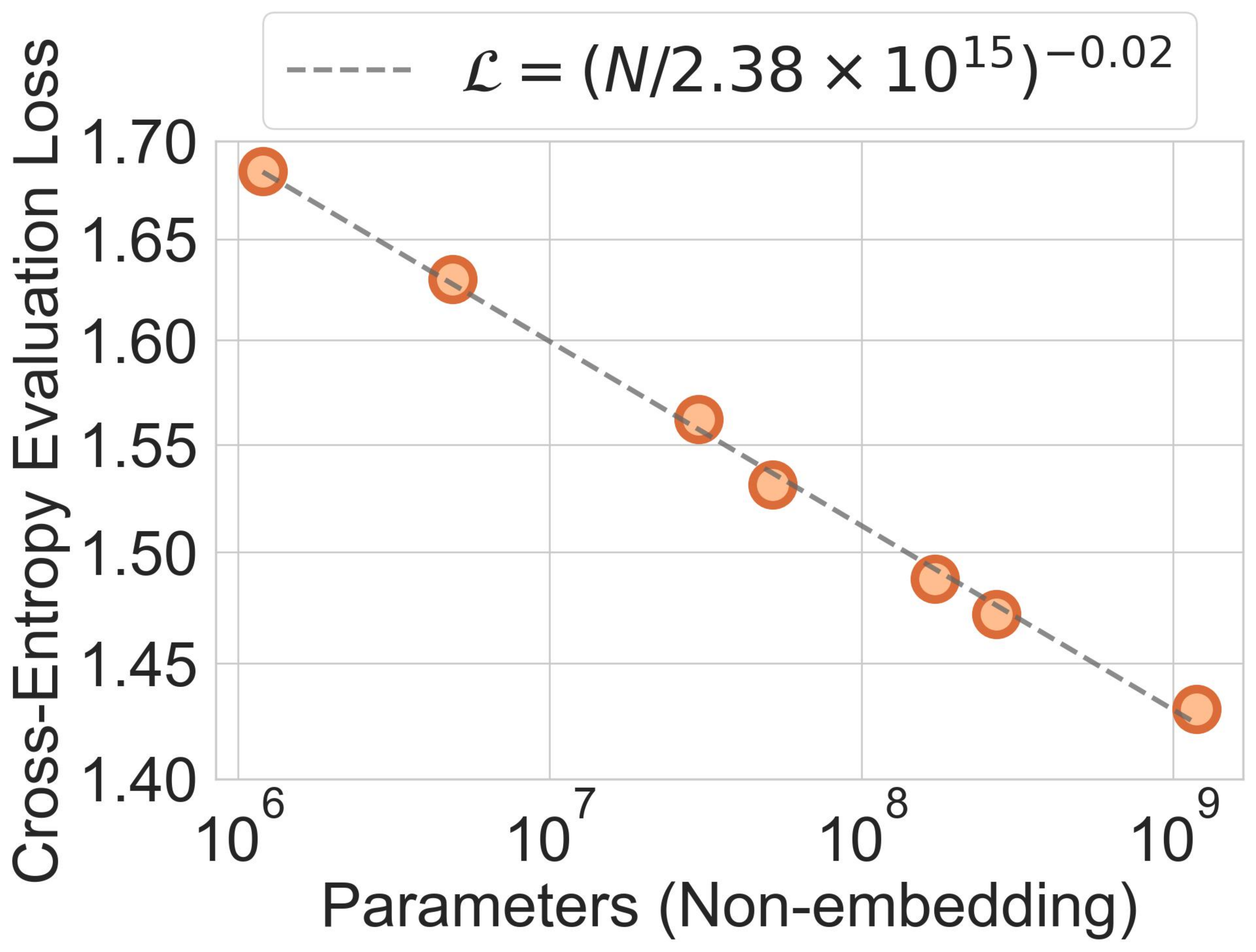}
    \caption{\textbf{Pre-Training Model Scaling Laws.} We pre-train models of varying sizes up to 1B parameters with other training hyperparameters held fixed on the Open Molecules 2025 4 million (4M) dataset. Evaluation performance improves in a clear power-law relationship with model size, with performance not saturating at the 1B parameter scale.}
    \label{fig:pt_model_scale} 
\end{figure}

We train seven Transformers of varying sizes up to one billion parameters on the OMol25 4M training split \citep{levine2025openmolecules2025omol25}. All runs use identical hyperparameters, with only the parameter count varied. We train models for 10 epochs with rotation augmentation, processing over 2B tokens, comparable to the ten billion token dataset used by Kaplan et al. \citep{kaplan2020scaling}.

Following prior work \citep{kaplan2020scaling, hoffmann2022trainingcomputeoptimallargelanguage}, we assume a power-law relationship between the validation loss $\mathcal{L}$ and model size $N$: $\mathcal{L}(N) = \left( \frac{N}{N_c} \right)^\alpha,$ where $N$ represents the number of non-embedding parameters. \Cref{fig:pt_model_scale} shows that Transformers continue to predictably improve in performance with model scale, with no sign of saturation up to 1B parameters.

\subsection{Uncertainty Quantification}
\label{sec:uq}
We examine the Transformer's ability to natively handle a broader set of tasks, beyond molecular property prediction, by evaluating it in a simple uncertainty quantification scenario. Since \methodname learn the whole joint distribution over tokens in its training set, our model can estimate the log probability of sequences. By looking at the log probability of an atomic structure predicted by our model, \methodname can be used as a tool for uncertainty quantification to identify structures that are out-of-distribution relative to the training dataset \citep{liu2024uncertaintyestimationquantificationllms, kuhn2023semanticuncertaintylinguisticinvariances, Abdar_2021}.

We use the SPICE dataset as a case study \citep{Eastman2023spice} and pre-train two Transformers (163M and 1B) to predict the energies and forces given atomic numbers and positions. We compare the predicted log probabilities from the two models to the force errors of commonly used ML architectures on the SPICE dataset \citep{gasteiger_gemnet_2021, kovács2023maceoff23, qu2024importance}.  

\begin{figure*}
    \centering
    \includegraphics[width=\linewidth]{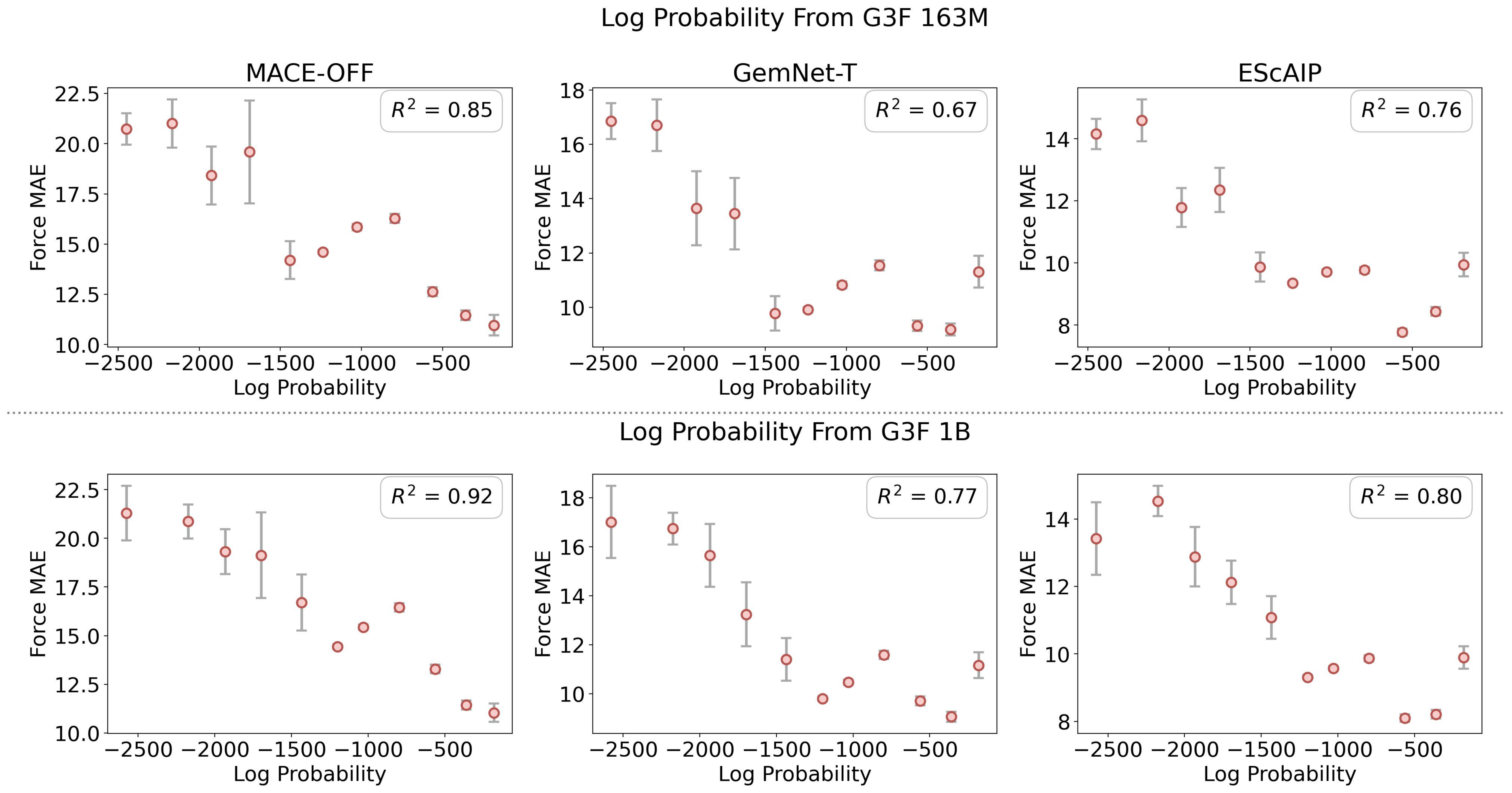}
    \caption{\textbf{\methodname can be used for uncertainty quantification for other ML models.} Since \methodname learns a joint distribution over positions, forces, and energies, it can compute log probabilities of new structures encountered at test-time. A higher log probability corresponds to the structure being closer to the training data. \methodname' predicted log probabilities correlate with force errors for other models like MACE-OFF, GemNet, and EScAIP. The correlation is stronger as \methodname is scaled up from 163M (top) to 1B parameters (bottom). Force errors are in meV/Å.}
    \label{fig:uq}
\end{figure*}

\Cref{fig:uq} shows that the Transformers' log probabilities are highly correlated with the force errors of MACE-OFF \citep{kovács2023maceoff23}, GemNet-T \citep{gasteiger_gemnet_2021}, and EScAIP \citep{qu2024importance}, indicating that it is accurately capturing the training distribution. As we scale the Transformer, it is able to better predict errors across all models. Although this is just a proof-of-concept, this previews how the flexibility of the Transformer can be used to easily tackle a broader class of problems, beyond just atomistic property prediction.   

\section{Inverse Distance Construction in the Attention Mechanism}
\label{apx:construction}

Supplementing the synthetic data experiments in the main text, we provide theoretical intuition for how Transformers can learn physically meaningful patterns in its attention weights. We provide details for our construction showing how the attention mechanism in a Transformer can approximate inverse-power-law relationships of the form $1 / r^{\alpha}$. This argument is not intended as a fully rigorous proof, but rather as an illustration that provides intuition for how Transformers can learn physically meaningful functions. Our presentation follows the style of prior work that uses explicit constructions to examine model behavior \citep{akyürek2023learningalgorithmincontextlearning}. For a more comprehensive treatment of the approximation power of Transformers, we refer the reader to \citep{yun2020transformersuniversalapproximatorssequencetosequence}.

\begin{prop}[Attention can represent $1/r^\alpha$ interactions]
\label{prop:attention_power_law}
Let $r_1, \ldots, r_n \in \mathbb{R}^3$ denote atomic Cartesian coordinates, and let
$r_{ij} = \|r_i - r_j\|_2$.
For any \(\alpha > 0\) and any compact interval $r_{ij} \in [r_{\min}, r_{\max}]$ with $r_{\min} > 0$,
there exists a Transformer attention layer with a finite number of heads such that the resulting attention output $A_{ij}$ satisfies $A_{ij}\;\approx\; \frac{1}{r_{ij}^{\alpha}}$.
\end{prop}

\begin{proof}[Construction]
We provide a construction for how the attention mechanism in a Transformer may be able to learn $1/r^\alpha$ relationships. Let each input embedding $x_i$ to the attention mechanism include Cartesian coordinates $r_i$, a constant, and the squared $L_2$ norm: $x_i = [r_i; 1, \|r_i\|_2^2]^\top \in \mathbb{R}^5$. The norm could be learned by the embedding layer, for instance, or previous attention layers. For an attention head $h$, we define linear query and key projections:
\begin{equation}
W_Q = -\sigma_h I_5, \quad W_K = \begin{bmatrix} 
-2 I_3 & \mathbf{0} & \mathbf{0} \\ 
\mathbf{0} & 0 & 1 \\ 
\mathbf{0} & 1 & 0 
\end{bmatrix}
\end{equation}
where $I_k$ is the $k \times k$ identity matrix. This results in queries and keys ($q_i = W_Q x_i$, $k_j = W_K x_j$) that yield attention logits proportional to the squared interatomic distance:
\begin{align}
q_i^\top k_j &= -\sigma_h \left( -2r_i \cdot r_j + \|r_j\|_2^2 + \|r_i\|_2^2 \right) \\
&= -\sigma_h \|r_i - r_j\|_2^2
\end{align}
After the softmax operation, the attention scores are proportional to $\exp(-\sigma_h r_{ij}^2)$, effectively forming Gaussian kernels. Given that Gaussian kernels are universal approximators \citep{gauss_universal}, the Transformer can approximate a $1/r^\alpha$ relationship by combining multiple such kernels across attention heads. Recalling the integral representation of the power function via the Gamma function, $\Gamma(z) = \int_0^\infty t^{z-1} e^{-t}dt$, we have:
\begin{equation}
\frac{1}{r^{\alpha}} = \frac{1}{\Gamma(\alpha/2)} \int_0^\infty t^{\alpha/2 - 1} e^{-r^2 t} dt
\end{equation}
For example, when $\alpha=2$, $\int_0^\infty e^{-r^2 t} dt = \frac{1}{r^2}$. The Transformer could approximate this integral as a discrete weighted sum across $H$ heads:
\begin{equation}
\frac{1}{r^\alpha} \approx \sum_{h=1}^{H} w_h e^{-\sigma_h r^2}
\end{equation}
By aggregating the outputs of these heads via the value projection and the final output dense layer, the model can learn the appropriate scaling factors $w_h$ and precision parameters $\sigma_h$ to reconstruct the desired potential.
\end{proof}

\section{Experiment Details}
\label{apx:experiment_details}

\subsection{Training} 

\paragraph*{Pre-training} We follow previous scaling laws literature when choosing hyperparameters for our pre-training experiments in \Cref{sec:md_and_sl} \citep{hoffmann2022trainingcomputeoptimallargelanguage, kaplan2020scaling}.  We use rotation augmentation during pre-training. We report our exact hyperparameters in \Cref{tab:hyperparameters} and our model sizes in \Cref{tab:model_sizes}.

We ran an ablation on the SPICE dataset (due to computational constraints) to evaluate the utility of adding the continuous input. With only the discretized input, the a 250M G3F model trained on the SPICE dataset has a force MAE of $\sim200$meV/Å after 5 epochs of training, whereas adding the continuous input brings this down to $\sim60$meV/Å. After the full pre-training, the fully discrete model has force errors of $\sim60$meV/Å versus $\sim40$mev/Å for the model with the continuous embeddings. 

\begin{figure*}[h]
    \centering
    \includegraphics[width=\linewidth]{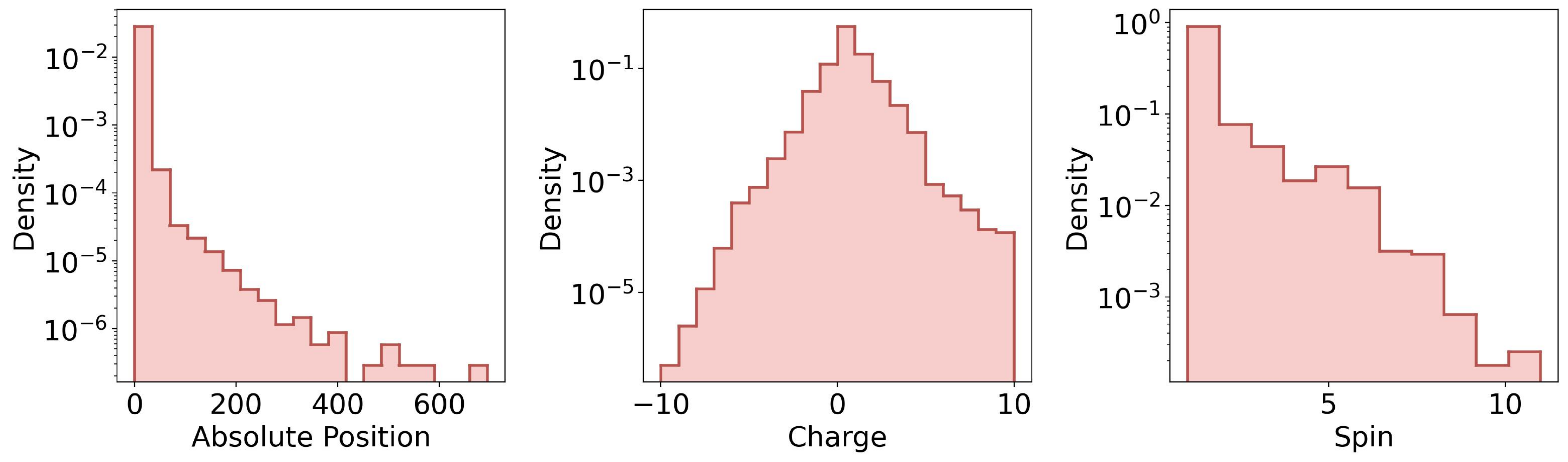}
    \caption{\textbf{Distribution of input features in the OMol25 dataset.} The continuous values in molecular datasets span multiple orders of magnitude and are heavy-tailed.}
    \label{fig:pos_dist}
\end{figure*}

We note that the continuous values present in molecular datasets are heavy tailed and span orders of magnitude (see \Cref{fig:pos_dist}). This causes large discretization errors at the tails with our quantile binning method. Giving the continuous sequence to the model as an additional input ameliorates this issue (see \Cref{sec:md_and_sl} and \Cref{fig:model_fig}). We think it is an interesting direction for future work to examine other discretization schemes for molecules.

\paragraph*{Fine-tuning} We also largely followed the same hyperparameters for fine-tuning; however, we found training to be more unstable during fine-tuning. Following previous large scale training recipes \citep{touvron2023llama2openfoundation, brown2020languagemodelsfewshotlearners}, we would resume training following an instability after halving the learning rate. We found it important to use a large enough batch size ($>1024$ structures) and clip the gradient norms around 100---not too high so as to destabilize training, but not too low so as to kill progress and get stuck in local minima. We also found it important to not use a discrete embedding for charge and spin, rather to treat it as a continuous signal to stabilize training, enabling gradients to propagate for these embeddings for any training sample. We hypothesize that this stabilized training due to the imbalance of extreme charge and spin in the dataset (see \Cref{fig:pos_dist}).  We also use rotation augmentation during fine-tuning.   We provide detailed hyperparameters in \Cref{tab:hyperparameters}.

\begin{figure*}[h]
    \centering
    \begin{lstlisting}[frame=single]
<BOS>

[POS] 
a_35: <NUM_579>
a_35: <NUM_728>
a_6: <NUM_657>
a_1: <NUM_456>
a_1: <NUM_766>
a_8: <NUM_240>
a_1: <NUM_231>
a_1: <NUM_340>
[POS_END]

[TARGET] <NUM_target_125> [TARGET_END]

[FORCE] 
<NUM_force_214> <NUM_force_35> <NUM_force_42> 
<NUM_force_75> <NUM_force_227> <NUM_force_72> 
<NUM_force_49> <NUM_force_210> <NUM_force_87> 
<NUM_force_209> <NUM_force_50> <NUM_force_197> 
<NUM_force_64> <NUM_force_94> <NUM_force_219> 
<NUM_force_229> <NUM_force_58> <NUM_force_210> 
<NUM_force_137> <NUM_force_206> <NUM_force_30> 
<NUM_force_25> <NUM_force_107> <NUM_force_174> 
[FORCE_END]

<EOS>
\end{lstlisting}
    
    \caption{\textbf{Example Discretized Input for the Model.} While we presented a simplified version of the input in the main text for clarity in \Cref{fig:disc_scheme} and \Cref{fig:model_fig}, we show a full example of a discretized input string here. We include special tokens to tell the model when to start predicting forces and energies (denoted by Target). a\_i represents atomic number $i$.}
    \label{fig:disc_input}
\end{figure*}

\begin{table*}[h]
    \centering
    \caption{\textbf{Hyperparameters for training Transformers.}}
    \begin{tabular}{lll}
\toprule
\textbf{Hyperparameter}                   & \textbf{Pre-Training Value}&Fine-Tuning Value\\ \hline
Learning Rate                    & $3 \times 10^{-4}$     &$3 \times 10^{-4}$\\
Weight Decay                     & 0.0                    &$1\times10^{-3}$\\
Optimizer                        & Adam                   &Adam\\
Epochs& 10&60\\
 Batch Size& 1024&2048\\
Warmup Percentage                & $5\%$                  &$10\%$\\
LR Scheduler                    & Cosine Decay           &Cosine Decay\\
Num Bins Force                   & 4096&-\\
Num Bins Target                  & 2048&-\\
Num Bins Pos (joint embedding)   & $10^3$                 &-\\
Num Bins Pos (1D discretization) & 512                    &-\\
 Output head(s)& Linear readout to logits&Energy + Force Gated MLPs\\
 Attention Mask& Causal&Bi-directional\\
 Loss& Cross-entropy&MAE (only E+F)\\
 Clip Grad Norm& 1.0&100\\
 \bottomrule
\end{tabular}

    \label{tab:hyperparameters}
\end{table*}

\begin{table*}[h]
    \centering
    \caption{\textbf{Model sizes for scaling experiments.} Model hyperparameters were adapted from \citet{hoffmann2022trainingcomputeoptimallargelanguage}.}
\begin{tabular}{ccccc}
\toprule
\textbf{\begin{tabular}[c]{@{}l@{}}Number of Non-Embedding \\ Parameters\end{tabular}} & \textbf{\begin{tabular}[c]{@{}l@{}}Hidden \\ Dimension\end{tabular}} & \textbf{\begin{tabular}[c]{@{}l@{}}Num\\ Layers\end{tabular}} & \textbf{\begin{tabular}[c]{@{}l@{}}Intermediate \\ Size\end{tabular}} & \textbf{\begin{tabular}[c]{@{}l@{}}Num \\ Heads\end{tabular}} \\ \hline
800k                                                                                   & 100                                                                  & 3                                                             & 400                                                                   & 4                                                             \\
5M                                                                                     & 256                                                                  & 4                                                             & 1024                                                                  & 4                                                             \\
 30M& 576& 5& 2304&9\\
50M                                                                                    & 576                                                                  & 9                                                             & 2304                                                                  & 9                                                             \\
 90M& 640& 13& 2560&10\\
170M                                                                                   & 768                                                                  & 18                                                            & 3072                                                                  & 12                                                            \\
250M                                                                                   & 1024                                                                 & 16                                                            & 4096                                                                  & 16                                                            \\
 350M& 1024& 20& 4096&16\\
1.2B                                                                                   & 1792                                                                 & 23                                                            & 7168                                                                  & 14                                                            \\ \bottomrule
\end{tabular}
    
    \label{tab:model_sizes}
\end{table*}

\section{Computational Details}
\label{apx: comp_details}

We trained our larger models for the scaling experiments on a cluster of  A100 and V100 GPUs. Training the 1B parameter model took $\sim750$ A100 hours using only data parallelism. We used gradient accumulation to achieve the effective batch size reported in \Cref{tab:hyperparameters}. We trained the 350M, 250M, and 170M parameter models for $\sim600$ V100 hours. The smaller models were trained on a single A6000 for up to 100 hours. We used compilation \citep{pytorch2} to speed up training. We think it is an interesting direction for future work to explore how systems innovations for the Transformer architecture can further speed up training and inference \citep{dao2022flashattentionfastmemoryefficientexact, pytorch2, kwon2023efficientmemorymanagementlarge}.

\end{document}